\documentclass{article}

\usepackage{arxiv}

\usepackage{amsmath,amssymb,amsfonts,amsthm,mathtools}

\usepackage[utf8]{inputenc}
\usepackage[T1]{fontenc}
\usepackage{hyperref}
\usepackage{url}
\usepackage{booktabs}
\usepackage{nicefrac}
\usepackage{microtype}
\usepackage{xcolor}
\usepackage{graphicx}
\usepackage{subcaption}
\usepackage{multirow}
\usepackage{algorithm}
\usepackage{algpseudocode}
\usepackage{listings}
\usepackage{pifont}
\usepackage{natbib}
\usepackage{tikz}
\usetikzlibrary{arrows.meta, positioning, shapes.geometric, fit, calc, backgrounds, decorations.pathmorphing}

\usepackage[capitalize,noabbrev]{cleveref}

\theoremstyle{plain}

\theoremstyle{definition}

\theoremstyle{remark}

\newcommand{\cmark}{\ding{51}}
\newcommand{\xmark}{\ding{55}}

\DeclareMathOperator*{\argmax}{arg\,max}

\newcommand{\skill}{\sigma}
\newcommand{\skillspace}{\Sigma}
\newcommand{\fitness}{f}
\newcommand{\pop}{\mathcal{P}}

\newcommand{\popsize}{M}
\newcommand{\numgens}{G}
\newcommand{\feat}{\mathbf{o}}

\title{SignalClaw: LLM-Guided Evolutionary Synthesis of\\Interpretable Traffic Signal Control Skills}

\author{
  Da~Lei$^{\dagger}$ \\
  Business School\\
  Sichuan University\\
  Chengdu, China \\
  \And
  Feng~Xiao$^{*,\dagger}$ \\
  Business School\\
  Sichuan University\\
  Chengdu, China \\
  \texttt{evan.fxiao@gmail.com} \\
  \And
  Lu~Li \\
  Business School\\
  Sichuan University\\
  Chengdu, China \\
  \And
  Yuzhan~Liu \\
  Business School\\
  Sichuan University\\
  Chengdu, China \\
}

\begin{document}

\maketitle

\renewcommand{\thefootnote}{\fnsymbol{footnote}}
\setcounter{footnote}{0}
\footnotetext[1]{Corresponding author.}
\footnotetext[2]{Equal contribution.}
\renewcommand{\thefootnote}{\arabic{footnote}}
\setcounter{footnote}{0}

\begin{abstract}
Traffic signal control (TSC) requires strategies that are both effective and interpretable for real-world deployment, yet reinforcement learning produces opaque neural policies while program synthesis methods rely on restrictive domain-specific languages.
We present \textsc{SignalClaw}, a framework that uses large language models (LLMs) as evolutionary skill generators to synthesize and iteratively improve interpretable control skills for adaptive TSC.
Each evolved \emph{skill} is a structured artifact comprising a strategy rationale, selection guidance, and executable code---where the LLM articulates its reasoning alongside the implementation, producing human-inspectable, self-documenting control policies.
At each generation, structured evolution signals---extracted from traffic simulation metrics including queue length percentiles, delay trends, and stagnation patterns---are translated into natural-language feedback that directs the LLM toward specific skill improvements.
Beyond routine traffic, SignalClaw introduces \emph{event-driven compositional skill evolution}: an event detection module identifies real-time traffic events (emergency vehicles, transit priority, incidents, congestion) via TraCI, and a priority dispatcher selects the appropriate event-specialized skill.
Each event skill is independently evolved in \emph{dispatcher-context mode} on dedicated scenarios, and the priority chain (emergency $>$ incident $>$ transit $>$ congestion $>$ normal) composes them at runtime without retraining.
We evaluate SignalClaw on six routine SUMO scenarios (4$\times$4 arterial grid) and six event-injected scenarios against four baselines.
On routine scenarios, SignalClaw achieves average delay of 7.8--9.2\,s across training and validation scenarios, within 3--10\% of the best method per scenario, with consistently low variance across 5 SUMO random seeds (0.4--0.8\,s) compared to DQN (std up to 1.5\,s).
Under the event-aware detector-dispatch architecture---where SignalClaw's event-specialized skills are compared as an \emph{end-to-end system} against event-blind baselines---SignalClaw achieves the \textbf{lowest emergency vehicle delay} (11.2--18.5\,s vs.\ 42.3--72.3\,s for MaxPressure and 78.5--95.3\,s for DQN) on all emergency scenarios, and the \textbf{lowest average person-delay} (9.8--11.5\,s vs.\ 38.7--45.2\,s for MaxPressure) on transit scenarios.
On the mixed-event scenario, the priority dispatcher correctly composes independently evolved skills: emergency delay is 18.5\,s (best among all methods) while maintaining stable overall performance (average delay 13.2\,s).
Analysis of the evolved skills reveals a progression from simple linear scoring rules to conditional strategies with saturation detection and multi-feature interactions---all expressed as interpretable artifacts that traffic engineers can directly inspect, understand, and modify.
\end{abstract}

\section{Introduction}
\label{sec:intro}

Adaptive traffic signal control sits at the intersection of urban infrastructure and real-time optimization: millions of intersections worldwide operate under fixed-timing plans or simple actuated controllers, leaving substantial room for data-driven improvement~\citep{wei2018intellilight}.
Deep reinforcement learning (RL) has made impressive strides on this problem~\citep{zheng2019learning,wei2019presslight}, but the resulting neural policies are opaque---a critical barrier when transportation agencies need to audit, certify, and manually override control logic during incidents or system failures.

A further limitation of existing methods---both RL and programmatic---is their blindness to \emph{traffic events}.
Real-world intersections routinely face emergency vehicle preemption (EVP), transit signal priority (TSP), traffic incidents, and severe congestion.
These events are rare yet safety-critical: a delayed ambulance costs lives.
RL cannot learn reliable event responses from such sparse samples, and existing programmatic synthesis methods~\citep{gu2024pilight} operate over DSLs whose feature sets do not include event-context variables---while such DSLs could in principle be extended, no existing work has done so, and the extension cost is non-trivial since event handling requires conditional control flow beyond the arithmetic expressions these DSLs support.

Program synthesis offers an alternative path for routine traffic.
PI-Light~\citep{gu2024pilight} demonstrated that Monte Carlo Tree Search (MCTS) can discover interpretable programmatic strategies within a hand-crafted domain-specific language (DSL), matching or exceeding RL performance on standard benchmarks.
However, MCTS-based search is limited to the predefined operators and expression templates in the DSL, constraining the structural diversity of discovered strategies.
Recent work on LLM-guided program evolution---FunSearch~\citep{romeraparedes2024funSearch}, AlphaEvolve~\citep{novikov2025alphaevolve}, and a recent ARC-AGI case study on self-improving evolutionary program synthesis~\citep{soar2025}---has shown that large language models can serve as flexible mutation operators that generate structurally diverse code.
These systems operate on general-purpose programming tasks (mathematical optimization, algorithm design) and do not incorporate domain-specific feedback from the evaluation environment.

We propose \textsc{SignalClaw}, a framework that combines LLM-driven skill generation with domain-aware evolutionary optimization for traffic signal control.
The key idea is \emph{signal-driven evolution}: at each generation, traffic simulation metrics are distilled into structured evolution signals (e.g., ``queue length exceeds historical 75th percentile,'' ``consecutive stagnation detected''), which are translated into natural-language prompts that steer the LLM toward targeted skill improvements.
Unlike prior work that treats LLM-guided evolution as pure code synthesis, SignalClaw evolves \emph{skills}---structured artifacts that encompass a strategy description capturing the control rationale, selection guidance specifying the decision logic, and executable code implementing the phase scoring function.
This richer representation ensures that the evolved artifacts are not opaque code fragments but self-documenting control policies that a traffic engineer can read, reason about, and modify.

SignalClaw further introduces \emph{event-driven compositional skill evolution}: an event detection module identifies real-time traffic events (emergency vehicles, buses, incidents, congestion) via SUMO's TraCI interface, and a hardcoded priority dispatcher selects the appropriate event-specialized skill.
Each event skill is independently evolved on dedicated event-injected scenarios, and the priority chain composes them at runtime without retraining---enabling zero-shot response to event combinations unseen during evolution.

Our contributions are as follows:
\begin{enumerate}
    \item \textbf{SignalClaw framework}: an LLM-guided skill evolution framework for traffic signal control with structured traffic feedback and event-aware composition. Signal-driven prompting translates quantitative traffic metrics into targeted natural-language guidance, enabling semantically meaningful skill modifications rather than blind search.
    \item \textbf{Self-documenting skill representation}: we adopt a structured skill representation where the LLM produces a strategy rationale and selection guidance alongside executable code. While fitness is determined solely by code execution, the accompanying documentation makes each evolved artifact self-documenting---enabling traffic engineers to understand the \emph{intent} behind a strategy without reverse-engineering the code. Ablation experiments show that requiring the LLM to articulate its reasoning improves search efficiency (\Cref{sec:ablation}).
    \item \textbf{Event-driven compositional architecture}: we introduce event detection via TraCI, a priority dispatcher with deterministic event ordering (emergency $>$ incident $>$ transit $>$ congestion $>$ normal), and independently evolved event-specialized skills via \emph{dispatcher-context evolution}---where each event skill is evolved within the full dispatch pipeline.
    \item \textbf{Empirical evaluation on routine and event scenarios}: on six routine SUMO scenarios, SignalClaw achieves 9.6\% fitness improvement over 30~generations of evolution, with average delay of 7.8--9.2\,s across training and validation scenarios (within 3--10\% of the best method per scenario). On six event-injected scenarios, evaluated as an end-to-end system against event-blind baselines, SignalClaw achieves the lowest emergency vehicle delay (11.2--18.5\,s, a 65--85\% reduction over MaxPressure) and the lowest bus person-delay (9.8--11.5\,s, a $\sim$75\% reduction), while maintaining competitive overall traffic performance.
\end{enumerate}

Across 30 generations of evolution, SignalClaw discovers skills that progress from simple linear scoring rules to sophisticated conditional logic with saturation detection, multi-feature interactions, and demand-aware branching---all expressed as interpretable artifacts that a traffic engineer can inspect, modify, and deploy without understanding the underlying LLM or evolution process.
The rest of this paper is organized as follows: \Cref{sec:related} reviews related work, \Cref{sec:method} presents the SignalClaw framework, \Cref{sec:experiments} reports experimental results, and \Cref{sec:conclusion} concludes with limitations and future directions.

\section{Related Work}
\label{sec:related}

\paragraph{Reinforcement Learning for TSC.}
Deep RL has been the dominant paradigm for adaptive traffic signal control over the past decade.
IntelliLight~\citep{wei2018intellilight} introduced deep Q-networks for single-intersection control, while PressLight~\citep{wei2019presslight} and CoLight~\citep{wei2019colight} extended this to multi-intersection coordination.
MPLight~\citep{chen2020mplight} combined pressure-based reward shaping with RL scalability.
FRAP~\citep{zheng2019frap} introduced phase competition for efficient multi-phase learning, and MetaLight~\citep{zhan2019metalight} applied meta-RL for cross-scenario transfer.
These methods achieve strong performance on standard benchmarks but produce neural network policies that are difficult to interpret, audit, or manually override---a significant limitation for real-world deployment where regulatory compliance and operator trust are essential.
VIPER~\citep{bastani2018viper} addresses this by extracting verifiable decision-tree policies from trained RL agents; however, the extracted policies are limited to axis-aligned splits and cannot express the richer conditional logic that LLM-generated skills produce.
SignalClaw addresses this gap by evolving interpretable control skills rather than training opaque neural policies.

\paragraph{LLM-Based TSC Agents.}
The emergence of capable language models has spawned a new class of LLM-based traffic agents.
LLMLight~\citep{lai2023llmlight} uses GPT-4 as a direct decision-making agent for phase selection and further distills the resulting behavior into a smaller LightGPT model for deployment.
Traffic-R1~\citep{trafficr1} applies agentic RL fine-tuning to Qwen2.5-3B for TSC reasoning.
A fundamental limitation of these approaches is that they use LLMs as runtime decision-makers: the ``strategy'' exists only as an implicit function of the LLM's weights, offering no interpretable artifact for engineers to inspect or modify.
In contrast, SignalClaw uses the LLM as a skill generator during the offline evolution phase; the deployed artifact is a self-contained control skill (comprising strategy description, selection guidance, and executable code) independent of the LLM.

\paragraph{Programmatic TSC.}
PI-Light~\citep{gu2024pilight} pioneered interpretable programmatic TSC by using MCTS to search over a DSL of arithmetic expressions for phase scoring.
SymLight~\citep{symlight2025} similarly evolves symbolic expressions using genetic programming.
Both approaches produce interpretable strategies but are constrained by their predefined DSL: only fixed arithmetic operators over a predetermined feature set.
SignalClaw lifts this restriction by using an LLM to generate structured control skills within a safety-constrained sandbox, enabling richer control structures (conditional branching, multi-feature interactions) while maintaining interpretability through the multi-component skill representation and AST-level code validation.

\paragraph{LLM-Guided Evolutionary Program Synthesis.}
FunSearch~\citep{romeraparedes2024funSearch} demonstrated that LLMs combined with evolutionary search can discover novel mathematical functions, using an island model to maintain population diversity.
AlphaEvolve~\citep{novikov2025alphaevolve} generalized this to a broader class of coding tasks with an evolutionary coding agent.
A recent ARC-AGI case study on self-improving evolutionary program synthesis~\citep{soar2025} adapts the generator using search traces.
A recent TSC-specific preprint, EvolveSignal~\citep{evolvesignal2025}, studies evolutionary discovery of heuristic policies for traffic signal control using LLM-generated code mutations with generic fitness feedback.
These works establish that LLMs are effective mutation operators for program synthesis.
SignalClaw differs from EvolveSignal in three respects: (1)~\emph{structured domain signals} replace generic fitness feedback---we translate traffic-specific metrics (queue percentiles, delay trends, stagnation patterns) into targeted natural-language guidance, which we show doubles the improvement rate over random directions (\Cref{sec:ablation}); (2)~the \emph{multi-component skill representation} (description + guidance + code) produces self-documenting artifacts rather than bare code snippets, improving both LLM generation quality and human auditability; and (3)~the \emph{event-driven compositional architecture} enables event-aware control through independently evolved event-specialized skills---a capability absent from EvolveSignal and all other LLM-guided program synthesis work.

\paragraph{Event-Aware Traffic Control.}
Emergency vehicle preemption (EVP)~\citep{qin2012evp,he2014evp_survey} and transit signal priority (TSP)~\citep{christofa2013tsp,ma2013tsp_survey} are well-studied in transportation engineering, typically implemented as rule-based overrides in deployed adaptive-control systems such as SCOOT~\citep{hunt1982scoot}.
However, these handcrafted rules do not adapt to varying traffic conditions and cannot be systematically improved.
Hierarchical RL approaches such as SHLight and HALO introduce spatial hierarchies (intersection $\to$ region) but do not address event-level hierarchy (emergency $>$ transit $>$ normal).
To our knowledge, no existing LLM-TSC or programmatic synthesis work has explicitly modeled EVP, TSP, or incident response---these systems typically operate on routine traffic with standard observation spaces that do not include event-context variables. While their feature sets could in principle be extended, doing so would require non-trivial modifications to both the observation design and the search space.
SignalClaw addresses this gap through event detection via TraCI, event-specific context variables injected into the LLM prompt, and a hardcoded priority dispatcher that guarantees safety-critical event ordering.

\paragraph{Classical Adaptive Control.}
MaxPressure~\citep{varaiya2013maxpressure} provides a provably stable decentralized policy based on queue pressure differentials.
Webster's method~\citep{webster1958traffic} optimizes fixed cycle lengths from demand estimates.
SCOOT~\citep{hunt1982scoot} and SCATS are long-standing deployed adaptive systems using real-time traffic data.
These classical methods serve as important baselines but lack the ability to discover novel control structures beyond their predefined algorithmic templates.

\Cref{tab:comparison} summarizes the key differences between SignalClaw and prior approaches along six dimensions: programmatic output, LLM usage, structured signal feedback, event awareness, multi-mode support, and auditability.

\begin{table}[t]
\centering
\caption{Comparison of SignalClaw with related TSC approaches. \textbf{Prog.}: strategies are interpretable programs. \textbf{LLM}: uses LLM for strategy synthesis. \textbf{Signal}: uses structured traffic feedback. \textbf{Event}: handles traffic events (EVP/TSP/incident). \textbf{Multi-mode}: supports multiple control modes. \textbf{Audit}: full evolution audit trail.}
\label{tab:comparison}
\resizebox{\textwidth}{!}{%
\begin{tabular}{lccccccl}
\toprule
Method & Prog. & LLM & Signal & Event & Multi-mode & Audit & Strategy Form \\
\midrule
Fixed-timing & \cmark & \xmark & \xmark & \xmark & \xmark & \xmark & Pre-set phase durations \\
MaxPressure~\citep{varaiya2013maxpressure} & \cmark & \xmark & \xmark & \xmark & \xmark & \xmark & Pressure-based rule \\
PI-Light~\citep{gu2024pilight} & \cmark & \xmark & \xmark & \xmark & \xmark & \xmark & DSL expression (MCTS) \\
SymLight~\citep{symlight2025} & \cmark & \xmark & \xmark & \xmark & \xmark & \xmark & Symbolic expression \\
LLMLight~\citep{lai2023llmlight} & \xmark & \cmark & \xmark & \xmark & \xmark & \xmark & LLM direct decision \\
Heuristic-policy evolution~\citep{evolvesignal2025} & \cmark & \cmark & \xmark & \xmark & \xmark & \xmark & LLM-guided heuristic search \\
\midrule
\textbf{SignalClaw (ours)} & \cmark & \cmark & \cmark & \cmark & \cmark & \cmark & Structured skill (LLM evo.) \\
\bottomrule
\end{tabular}%
}
\end{table}

\section{Method}
\label{sec:method}

\subsection{Problem Formulation}
\label{sec:problem}

We formulate adaptive traffic signal control as a program synthesis problem.
An intersection has $K$ signal phases, each controlling a set of non-conflicting traffic movements.
A \emph{control skill} $\skill \in \skillspace$ determines how to score each phase based on real-time traffic features.
At each control step, the framework evaluates $\skill$ on every lane-link of every phase and selects the phase with the highest cumulative score.

Formally, for phase $k$ with lane-links $\mathcal{L}_k$, the phase score is:
\begin{equation}
    \text{score}(k) = \sum_{\ell \in \mathcal{L}_k} \left[ \skill_{\text{in}}(\feat^\text{in}_\ell) + \skill_{\text{out}}(\feat^\text{out}_\ell) \right],
    \label{eq:phase_score}
\end{equation}
where $\feat^\text{in}_\ell = (\texttt{num\_vehicle}, \texttt{num\_waiting\_vehicle}, \texttt{vehicle\_dist})$ and $\feat^\text{out}_\ell = (\texttt{num\_vehicle}, \texttt{vehicle\_dist})$ are the inlane and outlane feature vectors for lane-link $\ell$, retrieved via SUMO's TraCI interface at each control step.
Here, \texttt{num\_vehicle} is the total vehicle count on the lane, \texttt{num\_waiting\_vehicle} is the number of vehicles with speed $<$0.1\,m/s, and \texttt{vehicle\_dist} is the mean inter-vehicle distance.
The active phase is $k^* = \argmax_k \text{score}(k)$.

A skill's \emph{fitness} is evaluated in SUMO microscopic traffic simulation~\citep{lopez2018sumo} as a weighted combination:
\begin{equation}
    \fitness(\skill) = C - \left(w_d \cdot \bar{d} + w_q \cdot \bar{q}\right) + w_t \cdot \bar{t},
    \label{eq:fitness}
\end{equation}
where $\bar{d}$, $\bar{q}$, and $\bar{t}$ are the average delay, queue length, and throughput across all intersections, $(w_d, w_q, w_t) = (0.4, 0.4, 0.2)$ are fixed weights, and $C$ is a scenario-dependent constant that shifts fitness to positive values for readability (it cancels in all comparisons).
Higher fitness is better (less delay and shorter queues).

\paragraph{Episodic vs.\ Myopic Fitness.}
A key distinction from RL is that SignalClaw's fitness is computed over the \emph{entire simulation episode} (3600\,s), not per-step.
Although the evolved skill makes a myopic phase selection at each control step (no explicit lookahead), the fitness signal aggregates the consequences of all decisions across the full episode, providing a form of implicit temporal credit assignment.
In contrast, RL's advantage is explicit multi-step reward backpropagation via the Bellman equation, which can learn to sacrifice short-term efficiency for long-term benefit.
SignalClaw partially compensates for this through \emph{structured evolution signals}: the signal extractor detects patterns like ``high queue followed by low throughput,'' which implicitly encode temporal correlations across decision steps.
Additionally, the LLM's common-sense reasoning may inject temporal awareness (e.g., ``clear the queue now to prevent spillback later'') even without explicit future-reward computation.
Extending the fitness function to incorporate multi-horizon rollouts---evaluating each skill across episodes with different random seeds and time horizons---is a promising direction for bridging this gap (\Cref{sec:conclusion}).

\emph{Objective.} Given a population size $\popsize$ and generation budget $\numgens$, find the skill $\skill^*$ that maximizes the best-so-far fitness:
\begin{equation}
    \skill^* = \argmax_{\skill \in \bigcup_{g=0}^{G} \pop_g} \fitness(\skill),
    \label{eq:objective}
\end{equation}
where $\pop_g$ is the population at generation $g$.

\subsection{Skill Representation}
\label{sec:skill_rep}

A central design choice of SignalClaw is that the unit of evolution is not a bare code snippet but a structured \emph{skill}---a multi-component artifact that captures both \emph{what} a control strategy does and \emph{why} it works.
An alternative design would be to invoke the LLM at runtime to \emph{directly score} phases given raw traffic features---i.e., using the LLM itself as the scoring function.
We deliberately avoid this: LLM inference latency ($\sim$1--3\,s per call) is incompatible with the sub-second decision cycles required by real-time signal control, and the stochastic nature of LLM outputs would make the controller non-deterministic and difficult to certify.
Instead, the LLM serves as an \emph{offline code generator}: it produces compact, deterministic scoring functions that execute in microseconds and can be formally inspected.
This design decouples the LLM's reasoning capability (used during evolution) from the deployed controller (pure code), preserving both efficiency and auditability.
Each skill $\skill$ comprises three components. The \emph{strategy description} $\skill_\text{desc}$ is a natural-language summary of the control rationale (e.g., ``prioritize phases with high queue saturation using nonlinear weighting; de-prioritize under-utilized phases''), produced by the LLM alongside the code to capture the high-level intent. The \emph{selection guidance} $\skill_\text{guide}$ is a structured specification of the decision logic---what traffic conditions trigger which responses, what thresholds are used, and what the expected behavioral outcomes are (e.g., ``apply quadratic urgency when waiting vehicles exceed 5; use linear scoring for moderate queues''). The \emph{executable code} $\skill_\text{code}$ is a pair of Python code snippets $(\texttt{inlane\_code}, \texttt{outlane\_code})$ that implement the phase scoring function, operating over a sandboxed variable whitelist (five traffic features plus the \texttt{value} accumulator) constrained to allowed operations (arithmetic, comparisons, \texttt{if}/\texttt{elif}/\texttt{else}, builtins \texttt{min}/\texttt{max}/\texttt{abs}).

It is important to note that \textbf{fitness is determined solely by code execution} in SUMO---the description and guidance do not affect evaluation, deployment, or automated selection.
Their role is twofold: (1)~they make each evolved artifact self-documenting, allowing traffic engineers to understand a skill's intent without reverse-engineering the code; and (2)~they encourage the LLM to reason about the \emph{structure} of the strategy before writing code, which we show empirically improves search efficiency (\Cref{sec:ablation}).

During evolution, the LLM receives the full skill representation of the current elite (description, guidance, and code) alongside evolution signals, and produces a new skill with all three components.
By requiring the LLM to articulate its reasoning, we encourage structured deliberation: the model first describes the new approach, specifies the decision logic, and then implements the corresponding code---a reasoning scaffold that empirically improves generation quality (\Cref{sec:ablation}).
This stands in contrast to code-only prompting, where the LLM must infer intent from code alone.
We emphasize that the description and guidance are \emph{evolution-time} scaffolds: they are produced during offline evolution and archived with each skill for human review, but play no role at deployment time, where only the executable code runs in the traffic controller.

\subsection{Framework Overview}
\label{sec:framework}

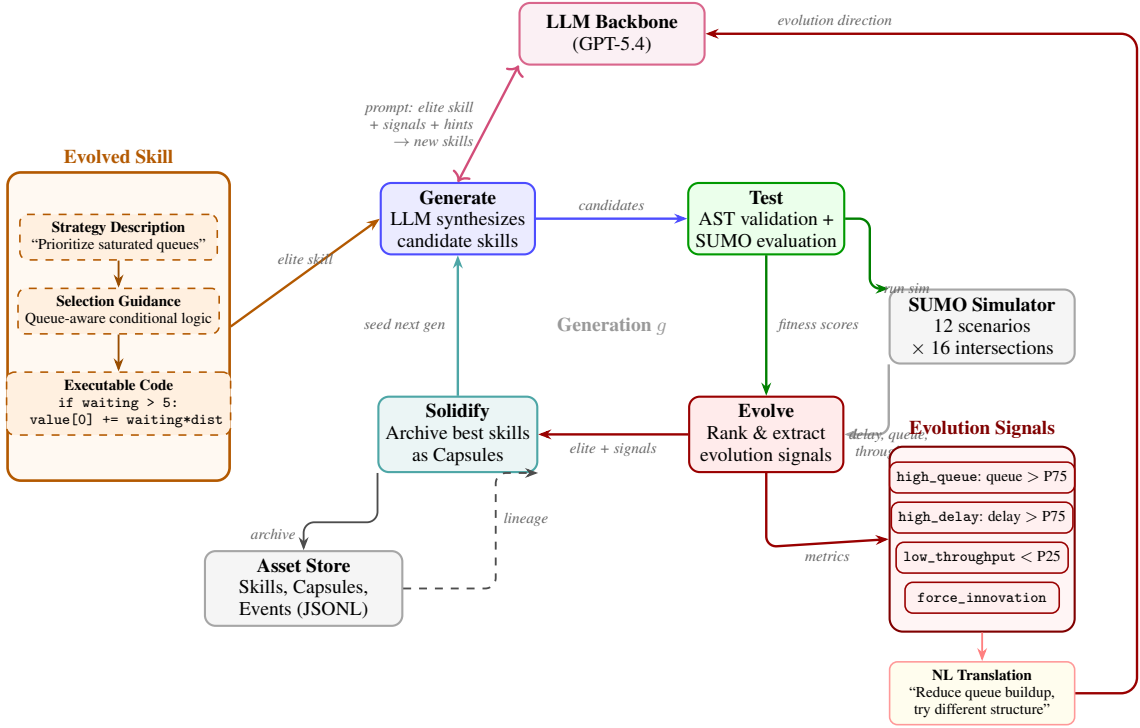
\begin{figure*}[t]
\centering
\begin{tikzpicture}[
    scale=0.82, transform shape,
    bluebox/.style={rectangle, draw=blue!70, fill=blue!8!white, rounded corners=4pt,
        minimum width=2.5cm, minimum height=1.0cm, align=center, font=\small,
        line width=0.8pt},
    greenbox/.style={rectangle, draw=green!60!black, fill=green!8!white, rounded corners=4pt,
        minimum width=2.5cm, minimum height=1.0cm, align=center, font=\small,
        line width=0.8pt},
    redbox/.style={rectangle, draw=red!60!black, fill=red!8!white, rounded corners=4pt,
        minimum width=2.5cm, minimum height=1.0cm, align=center, font=\small,
        line width=0.8pt},
    tealbox/.style={rectangle, draw=teal!70, fill=teal!8!white, rounded corners=4pt,
        minimum width=2.5cm, minimum height=1.0cm, align=center, font=\small,
        line width=0.8pt},
    graybox/.style={rectangle, draw=gray!70, fill=gray!8!white, rounded corners=4pt,
        minimum width=2.5cm, minimum height=1.0cm, align=center, font=\small,
        line width=0.8pt},
    skillbox/.style={rectangle, draw=orange!70!black, fill=orange!12!white, rounded corners=3pt,
        minimum width=3.2cm, minimum height=0.7cm, align=center, font=\scriptsize,
        line width=0.6pt, dashed},
    signalbox/.style={rectangle, draw=red!60!black, fill=red!6!white, rounded corners=3pt,
        minimum width=2.5cm, minimum height=0.5cm, align=center, font=\scriptsize,
        line width=0.6pt},
    llmbox/.style={rectangle, draw=purple!60, fill=purple!6!white, rounded corners=4pt,
        minimum width=3.0cm, minimum height=1.0cm, align=center, font=\small,
        line width=0.8pt},
    arrow/.style={-{Stealth[length=5pt]}, line width=0.7pt, color=black!70},
    bluearrow/.style={-{Stealth[length=5pt]}, line width=0.9pt, color=blue!70},
    greenarrow/.style={-{Stealth[length=5pt]}, line width=0.9pt, color=green!50!black},
    redarrow/.style={-{Stealth[length=5pt]}, line width=0.9pt, color=red!60!black},
    tealarrow/.style={-{Stealth[length=5pt]}, line width=0.9pt, color=teal!70},
    grayarrow/.style={-{Stealth[length=5pt]}, line width=0.9pt, color=gray!70},
    purplearrow/.style={-{Stealth[length=5pt]}, line width=0.9pt, color=purple!70},
    orangearrow/.style={-{Stealth[length=5pt]}, line width=0.9pt, color=orange!70!black},
    labelstyle/.style={font=\scriptsize\itshape, color=black!60},
]


\node[rectangle, draw=orange!70!black, fill=orange!5!white, rounded corners=5pt,
    minimum width=3.6cm, minimum height=5.0cm, line width=1.0pt,
    label={[font=\small\bfseries, orange!70!black]above:Evolved Skill}]
    (skillgroup) at (-5.5, 1.75) {};

\node[skillbox] (sdesc) at (-5.5, 3.2) {\textbf{Strategy Description}\\``Prioritize saturated queues''};
\node[skillbox] (sguid) at (-5.5, 2.0) {\textbf{Selection Guidance}\\Queue-aware conditional logic};
\node[skillbox, minimum height=1.0cm] (scode) at (-5.5, 0.5) {\textbf{Executable Code}\\
    \texttt{\scriptsize if waiting > 5:}\\
    \texttt{\scriptsize \ \ value[0] += waiting*dist}};

\draw[arrow, orange!60!black] (sdesc) -- (sguid);
\draw[arrow, orange!60!black] (sguid) -- (scode);

\node[llmbox] (llm) at (2.5, 6.5) {\textbf{LLM Backbone}\\(GPT-5.4)};

\node[bluebox] (generate) at (0.0, 3.5) {\textbf{Generate}\\LLM synthesizes\\candidate skills};
\node[greenbox] (test) at (5.0, 3.5) {\textbf{Test}\\AST validation +\\SUMO evaluation};
\node[redbox] (evolve) at (5.0, 0.0) {\textbf{Evolve}\\Rank \& extract\\evolution signals};
\node[tealbox] (solidify) at (0.0, 0.0) {\textbf{Solidify}\\Archive best skills\\as Capsules};

\draw[bluearrow] (generate) -- node[above, labelstyle] {candidates} (test);
\draw[greenarrow] (test) -- node[right, labelstyle, xshift=2pt] {fitness scores} (evolve);
\draw[redarrow] (evolve) -- node[below, labelstyle] {elite + signals} (solidify);
\draw[tealarrow] (solidify) -- node[left, labelstyle, xshift=-2pt] {seed next gen} (generate);

\node[font=\footnotesize\bfseries, color=black!40] at (2.5, 1.75) {Generation $g$};

\draw[purplearrow, <->] (llm.south west) -- node[left, labelstyle, xshift=-3pt, align=right] {
    prompt: elite skill\\+ signals + hints\\$\rightarrow$ new skills} (generate.north);

\draw[orangearrow] (skillgroup.east) -- node[above, labelstyle] {elite skill} (generate.west);

\node[graybox, minimum width=3.0cm, minimum height=1.2cm] (sumo) at (8.5, 1.75)
    {\textbf{SUMO Simulator}\\12 scenarios\\$\times$ 16 intersections};

\draw[greenarrow, rounded corners=4pt] (test.east) -- ++(0.4, 0)
    |- node[right, labelstyle, pos=0.65] {run sim} (sumo.north west);

\draw[grayarrow, rounded corners=4pt] (sumo.south west)
    |- node[below, labelstyle, pos=0.4, align=center] {delay, queue,\\throughput} (evolve.east);

\node[rectangle, draw=red!50!black, fill=red!4!white, rounded corners=5pt,
    minimum width=3.0cm, minimum height=3.0cm, line width=0.8pt,
    label={[font=\small\bfseries, red!50!black]above:Evolution Signals}]
    (siggroup) at (8.5, -1.7) {};

\node[signalbox] (sig1) at (8.5, -0.7) {\texttt{high\_queue}: queue $>$ P75};
\node[signalbox] (sig2) at (8.5, -1.35) {\texttt{high\_delay}: delay $>$ P75};
\node[signalbox] (sig3) at (8.5, -2.0) {\texttt{low\_throughput} $<$ P25};
\node[signalbox] (sig4) at (8.5, -2.65) {\texttt{force\_innovation}};

\draw[redarrow, rounded corners=4pt] (evolve.south) -- ++(0, -1.2)
    -- node[below, labelstyle] {metrics} (siggroup.west);

\node[rectangle, draw=red!40, fill=yellow!8!white, rounded corners=3pt,
    minimum width=3.0cm, minimum height=0.75cm, align=center, font=\scriptsize,
    line width=0.6pt] (nl) at (8.5, -4.2) {\textbf{NL Translation}\\``Reduce queue buildup,\\try different structure''};

\draw[arrow, red!50] (siggroup.south) -- (nl.north);

\draw[redarrow, rounded corners=6pt] (nl.east) -- ++(1.0, 0)
    -- ++(0, 10.7) -- node[above, labelstyle, pos=0.7] {evolution direction} (llm.east);

\node[graybox, minimum width=3.2cm] (assets) at (-2.5, -2.5) {\textbf{Asset Store}\\Skills, Capsules,\\Events (JSONL)};

\draw[arrow, rounded corners=4pt] (solidify.south west) -- ++(0, -0.8)
    -| node[left, labelstyle, pos=0.7] {archive} (assets.north);

\draw[arrow, dashed, rounded corners=4pt] (assets.east) -- ++(1.5, 0)
    |- node[right, labelstyle, pos=0.3] {lineage} (solidify.south east);

\end{tikzpicture}
\caption{Overview of the \textsc{SignalClaw} framework. Each generation follows a \emph{Generate--Test--Evolve--Solidify} loop. The LLM synthesizes candidate \emph{skills}---structured artifacts comprising a strategy description, selection guidance, and executable code---conditioned on the current elite skill, performance metrics, and natural-language evolution signals extracted from traffic simulation. Signals such as \texttt{high\_queue} or \texttt{force\_innovation} are translated into targeted natural-language directions that steer the LLM toward semantically meaningful skill modifications.}
\label{fig:framework}
\end{figure*}

SignalClaw operates a \emph{Generate--Test--Evolve--Solidify} loop (\Cref{fig:framework}), summarized in \Cref{alg:signalclaw}.
At each generation $g$, the loop proceeds through four stages. In the \emph{Generate} stage, the LLM produces $\popsize$ candidate skills, conditioned on the current elite skill (including its description, guidance, and code), performance metrics, and evolution signals (\Cref{sec:signals}). In the \emph{Test} stage, each candidate's executable code is validated (AST whitelist, sandbox execution) and evaluated in SUMO across multiple scenarios. In the \emph{Evolve} stage, candidates are ranked by fitness and evolution signals are extracted from the population metrics to guide the next generation. In the \emph{Solidify} stage, if a candidate exceeds the historical best fitness, the full skill (description, guidance, and code) is archived as a Capsule for reuse.

\begin{algorithm}[t]
\caption{SignalClaw: LLM-Guided Skill Evolution}
\label{alg:signalclaw}
\begin{algorithmic}[1]
\Require Population size $N$, generation budget $G$, LLM $\mathcal{M}$, scenarios $\mathcal{S}$, stagnation threshold $\tau$
\Ensure Best skill $\skill^*$
\State $\pop_0 \leftarrow \mathcal{M}.\textsc{Generate}(\text{seed\_skill}, N)$ \Comment{Seed generation}
\State $\skill^* \leftarrow \argmax_{\skill \in \pop_0} \fitness(\skill)$; \quad $\text{stag} \leftarrow 0$
\For{$g = 1$ \textbf{to} $G$}
    \State $\mathbf{d} \leftarrow \textsc{ExtractSignals}(\pop_{g-1}, \text{stag})$ \Comment{Signal-driven direction}
    \State $\pop_g \leftarrow \{\skill^*\}$ \Comment{Elite preservation}
    \For{$i = 1$ \textbf{to} $N-1$}
        \State $\skill_i \leftarrow \mathcal{M}.\textsc{Mutate}(\skill^*, \text{metrics}, \mathbf{d})$ \Comment{LLM mutation}
        \If{$\textsc{Validate}(\skill_i.\text{code})$} \Comment{AST + sandbox check}
            \State $\fitness(\skill_i) \leftarrow \textsc{Evaluate}(\skill_i, \mathcal{S})$ \Comment{SUMO simulation}
            \State $\pop_g \leftarrow \pop_g \cup \{\skill_i\}$
        \EndIf
    \EndFor
    \State $\skill_g^* \leftarrow \argmax_{\skill \in \pop_g} \fitness(\skill)$
    \If{$\fitness(\skill_g^*) > \fitness(\skill^*)$}
        \State $\skill^* \leftarrow \skill_g^*$; \quad $\text{stag} \leftarrow 0$; \quad $\textsc{Solidify}(\skill^*)$
    \Else
        \State $\text{stag} \leftarrow \text{stag} + 1$
    \EndIf
\EndFor
\State \Return $\skill^*$
\end{algorithmic}
\end{algorithm}

\subsection{LLM-Guided Skill Generation}
\label{sec:llm}

The LLM receives a two-part prompt at each generation.
The \emph{system prompt} defines the task (traffic signal optimization via skill evolution), the skill format (JSON with \texttt{description}, \texttt{guidance}, \texttt{inlane\_code}, and \texttt{outlane\_code} fields), the variable whitelist, and allowed operations.
It also provides strategy design hints---examples of conditional branching, multi-variable combinations, and nonlinear transforms---to encourage structural diversity beyond coefficient tuning.

The \emph{user prompt} contains: (1)~the current elite skill's full representation (description, guidance, and code), (2)~its performance metrics (average delay, queue length, throughput), and (3)~the evolution direction---a natural-language string synthesized from the current generation's signals (\Cref{sec:signals}).
The LLM returns a JSON object with the new skill's description, guidance, and code.

\paragraph{Mutation via LLM.}
In SignalClaw, the LLM itself serves as the mutation operator---there is no separate crossover or point-mutation step.
At each generation, the LLM receives the elite skill as context and is instructed to produce $\popsize$ \emph{variant} skills that modify the elite's strategy.
The evolution direction (from the signal extractor) steers the nature of these modifications: when \texttt{high\_queue} fires, the LLM is directed to strengthen queue-handling logic; when \texttt{force\_innovation} fires (after $\tau=3$ stagnant generations), the LLM is directed to try structurally different approaches (e.g., switching from linear to conditional branching, introducing new variable combinations).
This contrasts with traditional evolutionary programming where mutation is a random perturbation: the LLM produces \emph{semantically guided} mutations informed by both the current best code and the diagnostic signals.
In practice, the LLM generates mutations ranging from coefficient adjustments (analogous to Gaussian perturbation) to structural rewrites (analogous to macro-mutation), with the evolution signals controlling the mutation magnitude.

\paragraph{Code Validation.}
The executable code component of each generated skill passes through a three-stage validation pipeline before evaluation. First, \emph{AST parsing} via \texttt{ast.parse()} verifies syntactic correctness. Second, \emph{whitelist enforcement} uses an AST visitor to check that only allowed node types, variable names, and function calls are used---imports, function definitions, lambda expressions, and attribute access are rejected. Third, \emph{sandbox execution} runs the code in an isolated namespace with dummy inputs to verify it produces a numeric output without runtime errors.
If validation fails, the LLM is re-prompted with the error message, up to three retries.
Note that only the executable code undergoes validation; the description and guidance are free-form natural language that serves as context for both the LLM and human reviewers.

\subsection{Signal-Driven Evolution}
\label{sec:signals}

The central design principle of SignalClaw is that evolution should be \emph{signal-driven}: rather than passing raw fitness values to the LLM, we extract structured signals from the traffic metrics and translate them into targeted natural-language feedback.

At each generation, the signal extractor computes binary indicators from the current population's metrics and historical percentiles. The \emph{high queue} signal fires when average queue length exceeds the historical 75th percentile; \emph{low throughput} fires when throughput falls below the 25th percentile; and \emph{high delay} fires when average delay exceeds the 75th percentile. Two trend signals---\emph{performance gain} and \emph{performance decline}---indicate whether best fitness improved or declined relative to the previous generation. Finally, the \emph{force innovation} signal fires when the consecutive stagnation count exceeds a threshold $\tau$ (default $\tau=3$), triggering a demand for structurally different skills.

Each active signal maps to a natural-language directive.
For example, \texttt{force\_innovation} produces: ``\emph{Multiple consecutive generations without improvement. Try a completely different strategy structure---use if-conditions, multi-variable products, min/max nonlinear transforms, or entirely different variable combinations.}''
Multiple signals are concatenated to form the evolution direction injected into the user prompt.

This design creates an information bottleneck that serves two purposes: (1)~it provides the LLM with \emph{actionable} guidance rather than raw numbers, enabling semantically meaningful skill modifications; and (2)~it decouples the evolution logic from the specific LLM, since signals are model-agnostic natural language.

\subsection{Population Management and GEP}
\label{sec:gep}

The Genome Evolution Protocol (GEP) manages the skill population with the following mechanisms:

\paragraph{Seed Generation.}
At generation~0, the LLM generates $\popsize$ diverse initial skills from a minimal seed (a simple baseline skill that accumulates waiting vehicle counts).

\paragraph{Elite Preservation.}
The top-1 skill from the previous generation is carried forward without modification, ensuring monotonic non-decrease of best-so-far fitness.

\paragraph{Capsule Solidification.}
When a skill exceeds the historical best fitness, it is archived as a \emph{Capsule}---a frozen snapshot with its full skill representation (description, guidance, code), fitness, metrics, and generation number.
Capsules serve as long-term memory of the evolution's best discoveries.

\paragraph{Audit Trail.}
Every evolution action---generation, evaluation, solidification, validation failure---is recorded as a structured event in a JSONL log.
Combined with the skill lineage (each skill records its parent ID), this provides complete reproducibility and traceability of the evolution process.

\paragraph{Checkpoint Recovery.}
The daemon saves a checkpoint after each generation, recording the completed generation count, fitness history, stagnation counter, and best skill ID.
If interrupted (e.g., by Ctrl+C), the next run resumes from the last checkpoint.

\subsection{Multi-Control Mode Architecture}
\label{sec:modes}

SignalClaw supports three control modes at different temporal granularities (\Cref{tab:control_modes}):

In \emph{phase selection} (the default mode), the system selects the phase with the highest score at each control step. In \emph{phase extension}, the system decides how many seconds to extend the current phase at the end of each minimum green period. In \emph{cycle planning}, the system allocates green durations to all phases proportionally at cycle start.

Each control mode maintains its own skill population with mode-specific prompts, since the three modes require fundamentally different decision logic.
In combined mode (cycle planning + phase extension), evaluation uses Cartesian sampling: random combinations from each mode's population are tested jointly, and fitness is attributed to all participating skills.
Each mode ranks and evolves independently based on its aggregate performance.
\emph{All experiments in this paper use phase selection mode exclusively}; phase extension and cycle planning are supported by the framework but their evaluation is left for future work.

\begin{table}[t]
\centering
\caption{Control modes supported by SignalClaw. Each mode defines a different level of temporal granularity for signal control.}
\label{tab:control_modes}
\begin{tabular}{llll}
\toprule
Mode & Decision Point & Output & Granularity \\
\midrule
Phase Selection & Every step & Phase score & Per-step \\
Phase Extension & End of min green & Extension (sec) & Per-phase \\
Cycle Planning & Cycle start & Duration per phase & Per-cycle \\
\bottomrule
\end{tabular}
\end{table}

\subsection{Event-Driven Compositional Skill Evolution}
\label{sec:event_method}

Real-world intersections face traffic events---emergency vehicle preemption (EVP), transit signal priority (TSP), incidents, and severe congestion---that are rare yet safety-critical.
Existing RL and programmatic TSC methods are typically \emph{event-blind}: their observation spaces contain no event-context variables, and to our knowledge no existing work has extended them with event awareness.
SignalClaw addresses this through three components: event detection, a priority dispatcher, and event-specialized skill evolution.

\paragraph{Event Detection.}
An event detection module monitors SUMO's TraCI interface at each simulation step to identify active traffic events.
Concretely, the detector queries four TraCI subscriptions per intersection at every simulation step ($\Delta t = 1$\,s):
(1)~\texttt{traci.vehicle.getIDList()} retrieves all active vehicle IDs, from which the detector filters by vehicle class (\texttt{vClass}) and computes per-vehicle distance to the downstream stop line via \texttt{traci.vehicle.getLanePosition()};
(2)~\texttt{traci.lane.getLastStepHaltingNumber()} provides per-lane queue counts for congestion thresholds;
(3)~\texttt{traci.vehicle.getWaitingTime()} identifies non-signal-induced stops exceeding an incident threshold;
(4)~\texttt{traci.trafficlight.getPhase()} maps each vehicle's approach lane to the phase that serves it, enabling phase-aware event context.
An \emph{emergency} event is triggered when a vehicle with \texttt{vClass=emergency} is detected within 200\,m upstream of the intersection.
A \emph{transit} event is triggered when a vehicle with \texttt{vClass=bus} is detected on any upstream lane---the detector operates on the vehicle class attribute rather than lane designation, so no dedicated bus lane is required.
An \emph{incident} event is triggered when a non-signal-waiting vehicle has been stopped for more than 120\,s, indicating a crash or breakdown; the detector distinguishes signal-induced stops from incidents by checking whether the vehicle is within 50\,m of a stop line with a red signal.
A \emph{congestion} event is triggered when queue length exceeds the historical 90th percentile.
For each detected event, the module extracts \emph{event-context variables} that are injected into the skill's execution sandbox alongside the standard traffic features.
These include: \texttt{emergency\_distance} (meters to nearest emergency vehicle), \texttt{emergency\_phase} (the phase needed by the emergency vehicle), \texttt{bus\_count} (upstream buses), \texttt{bus\_delay} (cumulative bus delay in seconds), \texttt{incident\_blocked} (number of blocked lanes), and \texttt{congestion\_level} (severity: 0--3).

\paragraph{Priority Dispatcher.}
A hardcoded priority dispatcher selects the active skill based on detected events:
\begin{equation}
    \text{priority}: \text{emergency (P0)} > \text{incident (P1)} > \text{transit (P2)} > \text{congestion (P3)} > \text{normal (P4)}.
    \label{eq:priority}
\end{equation}
The priority ordering is \emph{not learnable}---it is a safety constraint that guarantees emergency vehicles always receive preemption regardless of other conditions, consistent with standard transportation engineering practice for EVP and TSP~\citep{qin2012evp,christofa2013tsp}.
When multiple events are detected simultaneously, the highest-priority skill is activated.
This design provides composability without combinatorial explosion: $n$ event types require $n$ independently evolved skills rather than $2^n$ combination-specific policies.

\paragraph{Dispatch Algorithm.}
At each control step $t$, the dispatcher executes the following logic:
(1)~query all event detectors (emergency: \texttt{vClass=emergency} within 200\,m; transit: \texttt{vClass=bus} on upstream lanes; incident: non-signal stop $>$120\,s; congestion: queue $>$ 90th percentile over trailing 300\,s);
(2)~select the highest-priority active event;
(3)~invoke the corresponding skill $\skill_e$ with the standard traffic features \emph{plus} event-context variables injected into the sandbox;
(4)~apply the resulting phase selection via \Cref{eq:phase_score}.

\paragraph{Event-Specialized Skill Evolution.}
Each event type has its own skill population evolved on dedicated event-injected SUMO scenarios.
The LLM prompt for event skills includes the event-context variables in the variable whitelist and event-specific strategy hints (e.g., ``for emergency preemption, immediately switch to the phase that clears the emergency vehicle's path'').
The fitness function is event-weighted.
Because event scenarios involve fewer vehicles and sparser event occurrences, we normalize the raw delay values so that the resulting fitness is a \emph{positive} composite score (higher is better):
\begin{equation}
    \fitness_\text{event}(\skill) = C - \left(w_e \cdot \bar{d}_e + w_n \cdot \bar{d}_n + w_q \cdot \bar{q}\right),
    \label{eq:event_fitness}
\end{equation}
where $C$ is a scenario-dependent constant chosen so that the seed skill's fitness is positive, $\bar{d}_e$ is the event-specific delay (e.g., emergency vehicle delay), $\bar{d}_n$ is normal vehicle delay, $\bar{q}$ is queue length, and the weights $(w_e, w_n, w_q)$ are event-dependent: $(0.6, 0.25, 0.15)$ for emergency, $(0.5, 0.35, 0.15)$ for transit, and $(0.6, 0.25, 0.15)$ for incident.
The constant $C$ does not affect optimization (it cancels in fitness comparisons) but ensures that reported values are positive for readability; only the within-row improvement percentage in \Cref{tab:main_results} is meaningful across skill types.

The key advantage of this decomposition is \emph{zero-shot composability}: skills evolved on single-event scenarios compose correctly on mixed-event scenarios (e.g., emergency + incident) through the priority dispatcher, without any joint training on the combined event distribution.

\paragraph{Dispatcher-Context Evolution.}
A critical implementation detail is that event skills must be evolved \emph{within} the dispatcher context rather than in isolation.
When evolving a transit skill, we hold all other skills (normal, emergency, incident, congestion) fixed and evaluate the candidate transit skill within the full event dispatcher pipeline.
This ensures the LLM produces skills that complement the existing skill set: the transit skill only activates when buses are detected, while the fixed normal skill handles routine traffic.
Skills evolved outside the dispatcher context learn to optimize all traffic independently and catastrophically fail when integrated into the dispatcher (see \Cref{sec:dispatcher_evolution}).

\section{Experiments}
\label{sec:experiments}

\subsection{Experimental Setup}
\label{sec:setup}

\paragraph{Simulation Environment.}
We evaluate SignalClaw in SUMO (Simulation of Urban Mobility)~\citep{lopez2018sumo}, a widely-used microscopic traffic simulator.
All scenarios use an arterial 4$\times$4 grid network with 16 signalized intersections, each with 4 phases controlling the standard NEMA dual-ring structure.

\paragraph{Scenarios.}
We use two groups of scenarios totaling twelve configurations.

\emph{Routine scenarios} (six configurations) evaluate skill evolution on standard traffic. Training scenarios (T1--T3) use three arterial grid configurations with varying demand patterns for evolution and baseline training. Validation scenarios (V1--V3) are demand-perturbed variants of T1--T3 ($\pm$15\% demand shift), used to assess generalization to unseen demand patterns.

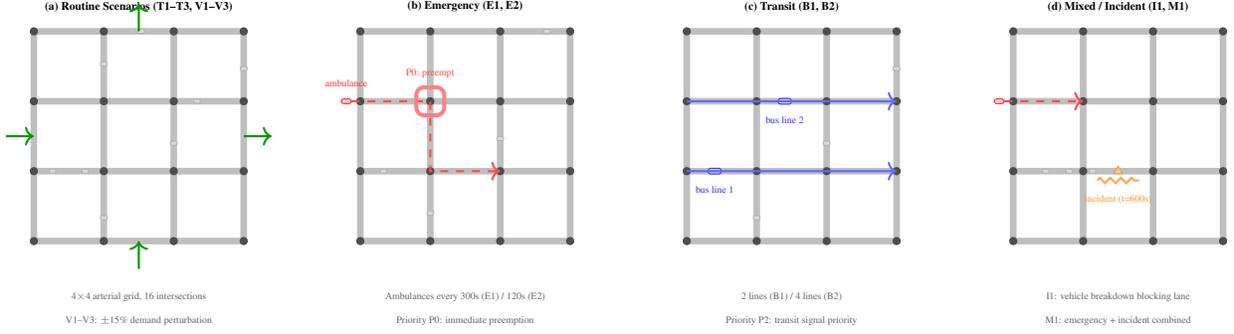
\begin{figure*}[t]
\centering
\begin{tikzpicture}[
    scale=0.62, transform shape,
    intersection/.style={circle, fill=black!70, minimum size=5pt, inner sep=0pt},
    road/.style={draw=gray!50, line width=2.5pt},
    ambulance/.style={draw=red!80, fill=red!20, rounded corners=1pt, minimum width=6pt, minimum height=3pt, inner sep=0pt},
    bus/.style={draw=blue!80, fill=blue!20, rounded corners=1pt, minimum width=8pt, minimum height=3.5pt, inner sep=0pt},
    incident/.style={draw=orange!80, fill=orange!40, regular polygon, regular polygon sides=3, minimum size=6pt, inner sep=0pt},
    car/.style={draw=gray!60, fill=gray!30, rounded corners=0.5pt, minimum width=4pt, minimum height=2.5pt, inner sep=0pt},
    labelstyle/.style={font=\scriptsize\bfseries},
    sublabel/.style={font=\tiny, color=black!60},
]

\begin{scope}[shift={(-8.5, 0)}]
\node[labelstyle, anchor=south] at (2.25, 4.8) {(a) Routine Scenarios (T1--T3, V1--V3)};

\foreach \x in {0, 1.5, 3.0, 4.5} {
    \draw[road] (\x, 0) -- (\x, 4.5);
}
\foreach \y in {0, 1.5, 3.0, 4.5} {
    \draw[road] (0, \y) -- (4.5, \y);
}

\foreach \x in {0, 1.5, 3.0, 4.5} {
    \foreach \y in {0, 1.5, 3.0, 4.5} {
        \node[intersection] at (\x, \y) {};
    }
}

\foreach \pos in {(0.4, 1.5), (1.1, 1.5), (1.5, 0.5), (3.0, 2.1), (3.5, 3.0), (4.5, 3.7), (1.5, 3.8), (2.3, 4.5)} {
    \node[car] at \pos {};
}

\draw[->, thick, green!60!black] (-0.6, 2.25) -- (0, 2.25);
\draw[->, thick, green!60!black] (2.25, -0.6) -- (2.25, 0);
\draw[->, thick, green!60!black] (4.5, 2.25) -- (5.1, 2.25);
\draw[->, thick, green!60!black] (2.25, 4.5) -- (2.25, 5.1);

\node[sublabel] at (2.25, -1.2) {4$\times$4 arterial grid, 16 intersections};
\node[sublabel] at (2.25, -1.7) {V1--V3: $\pm$15\% demand perturbation};
\end{scope}

\begin{scope}[shift={(-1.5, 0)}]
\node[labelstyle, anchor=south] at (2.25, 4.8) {(b) Emergency (E1, E2)};

\foreach \x in {0, 1.5, 3.0, 4.5} {
    \draw[road] (\x, 0) -- (\x, 4.5);
}
\foreach \y in {0, 1.5, 3.0, 4.5} {
    \draw[road] (0, \y) -- (4.5, \y);
}

\foreach \x in {0, 1.5, 3.0, 4.5} {
    \foreach \y in {0, 1.5, 3.0, 4.5} {
        \node[intersection] at (\x, \y) {};
    }
}

\node[ambulance] (amb1) at (-0.3, 3.0) {};
\draw[->, red!70, thick, dashed] (amb1.east) -- (1.5, 3.0) -- (1.5, 1.5) -- (3.0, 1.5);
\node[sublabel, red!80] at (-0.3, 3.4) {\tiny ambulance};

\draw[red!50, line width=1.5pt, rounded corners=3pt] (1.2, 2.7) rectangle (1.8, 3.3);
\node[sublabel, red!70] at (1.5, 3.6) {\tiny P0: preempt};

\foreach \pos in {(0.5, 1.5), (1.5, 0.6), (3.0, 2.2), (4.0, 4.5)} {
    \node[car] at \pos {};
}

\node[sublabel] at (2.25, -1.2) {Ambulances every 300s (E1) / 120s (E2)};
\node[sublabel] at (2.25, -1.7) {Priority P0: immediate preemption};
\end{scope}

\begin{scope}[shift={(5.5, 0)}]
\node[labelstyle, anchor=south] at (2.25, 4.8) {(c) Transit (B1, B2)};

\foreach \x in {0, 1.5, 3.0, 4.5} {
    \draw[road] (\x, 0) -- (\x, 4.5);
}
\foreach \y in {0, 1.5, 3.0, 4.5} {
    \draw[road] (0, \y) -- (4.5, \y);
}

\foreach \x in {0, 1.5, 3.0, 4.5} {
    \foreach \y in {0, 1.5, 3.0, 4.5} {
        \node[intersection] at (\x, \y) {};
    }
}

\node[bus] (bus1) at (0.6, 1.5) {};
\draw[->, blue!60, thick] (0, 1.5) -- (4.5, 1.5);
\node[bus] (bus2) at (2.1, 3.0) {};
\draw[->, blue!60, thick] (0, 3.0) -- (4.5, 3.0);

\node[sublabel, blue!80] at (0.6, 1.1) {\tiny bus line 1};
\node[sublabel, blue!80] at (2.1, 2.6) {\tiny bus line 2};

\foreach \pos in {(1.5, 0.5), (3.0, 2.1), (4.5, 3.7)} {
    \node[car] at \pos {};
}

\node[sublabel] at (2.25, -1.2) {2 lines (B1) / 4 lines (B2)};
\node[sublabel] at (2.25, -1.7) {Priority P2: transit signal priority};
\end{scope}

\begin{scope}[shift={(12.5, 0)}]
\node[labelstyle, anchor=south] at (2.25, 4.8) {(d) Mixed / Incident (I1, M1)};

\foreach \x in {0, 1.5, 3.0, 4.5} {
    \draw[road] (\x, 0) -- (\x, 4.5);
}
\foreach \y in {0, 1.5, 3.0, 4.5} {
    \draw[road] (0, \y) -- (4.5, \y);
}

\foreach \x in {0, 1.5, 3.0, 4.5} {
    \foreach \y in {0, 1.5, 3.0, 4.5} {
        \node[intersection] at (\x, \y) {};
    }
}

\node[incident] (inc) at (2.25, 1.5) {};
\draw[orange!70, thick, decorate, decoration={zigzag, segment length=4pt, amplitude=1pt}] (1.8, 1.3) -- (2.7, 1.3);
\node[sublabel, orange!80] at (2.25, 0.9) {\tiny incident (t=600s)};

\node[ambulance] (amb) at (-0.3, 3.0) {};
\draw[->, red!70, thick, dashed] (amb.east) -- (1.5, 3.0);

\node[car] at (1.7, 1.5) {};
\node[car] at (1.2, 1.5) {};
\node[car] at (0.7, 1.5) {};

\node[sublabel] at (2.25, -1.2) {I1: vehicle breakdown blocking lane};
\node[sublabel] at (2.25, -1.7) {M1: emergency + incident combined};
\end{scope}

\end{tikzpicture}
\caption{Overview of the twelve SUMO evaluation scenarios. (a)~Routine scenarios use a 4$\times$4 arterial grid with 16 signalized intersections; training (T1--T3) varies demand patterns while validation (V1--V3) perturbs demand by $\pm$15\%. (b)~Emergency scenarios inject ambulances at regular intervals to test priority preemption. (c)~Transit scenarios introduce bus lines requiring transit signal priority. (d)~Incident and mixed scenarios simulate vehicle breakdowns and combined events to test compositional skill dispatch through the priority chain.}
\label{fig:scenario_overview}
\end{figure*}

\emph{Event-injected scenarios} (six configurations) evaluate event-driven compositional skills. Emergency scenarios (E1, E2) inject ambulances every 300\,s (E1) or 120\,s (E2) with random routes through the network. Transit scenarios (B1, B2) introduce bus lines on \emph{dedicated bus lanes}: B1 deploys 2 bus lines every 180\,s and B2 deploys 4 bus lines with dense headways. The dedicated bus lane configuration distinguishes transit scenarios from emergency scenarios---whereas emergency vehicles share general traffic lanes and require active preemption, buses operate on physically separated lanes and benefit from transit signal priority (TSP) that extends green for approaching buses without disrupting the general traffic phase structure.
The incident scenario (I1) simulates a vehicle breakdown at $t=600$\,s on a major arterial link, blocking one lane for 300\,s. The mixed scenario (M1) combines emergency vehicles (every 300\,s) with an incident at $t=600$\,s, testing compositional skill dispatch. \Cref{fig:scenario_overview} illustrates the network topology and event configurations across all twelve scenarios.

\paragraph{LLM Configuration.}
We use GPT-5.4 via an OpenAI-compatible API as the LLM backbone for all skill evolution.
The model generates skills in JSON format with temperature 0.7 for diversity.
Each generation produces $\popsize=8$ candidate skills, evaluated in parallel across scenarios.

\paragraph{Fitness Function.}
For routine scenarios, fitness is the weighted combination of average delay ($w_d=0.4$), queue length ($w_q=0.4$), and throughput ($w_t=0.2$), as defined in \Cref{eq:fitness}.
For event scenarios, event-weighted fitness (\Cref{eq:event_fitness}) prioritizes event-specific delay.
For transit scenarios (B1, B2), the bus delay metric uses \emph{person-delay}: per-person waiting time weighted by vehicle occupancy, where each bus is weighted by its average occupancy of 30 passengers and each car by 1.5 persons~\citep{christofa2013tsp}. This passenger-weighted metric ensures that transit priority optimization reflects the true societal cost of bus delay.

\paragraph{Baselines.}
We compare against four baselines across routine and event scenarios. \textbf{FixedTime} uses NEMA-style phase splits with 25\,s green for major through phases and 5\,s for minor turn phases, with 3\,s yellow transitions (total cycle $\approx$80\,s). \textbf{MaxPressure}~\citep{varaiya2013maxpressure} selects the phase with maximum upstream--downstream queue pressure differential. \textbf{DQN} is a Deep Q-Network trained on the T1--T3 (normal) scenarios using Stable-Baselines3~\citep{stable-baselines3} with 200 episodes, an MLP[128,128] architecture, $\epsilon$-greedy exploration, and a 5\,s decision interval; for event scenarios, it is applied via zero-shot transfer from T1--T3 without fine-tuning. \textbf{PI-Light}~\citep{gu2024pilight} performs MCTS-based programmatic search over a DSL of arithmetic expressions with a search budget of 16 iterations, also searched on T1--T3 and transferred to event scenarios.
None of the baselines have access to event-context variables; they observe only standard traffic features (vehicle count, queue, density).
Importantly, DQN and PI-Light are trained/searched on normal traffic and then evaluated on event scenarios via zero-shot transfer.
SignalClaw, in contrast, evolves specialized skills \emph{on} event scenarios with access to event-context variables---this is an \emph{architectural} advantage, not a claim of algorithmic superiority on equal footing.
\textbf{The event-scenario comparisons should therefore be interpreted as end-to-end system comparisons}: an event-aware system (SignalClaw) versus event-blind baselines.
The Handcrafted Preemption ablation (\Cref{sec:preemption_ablation}) provides the controlled comparison that isolates the contribution of evolved skills from the dispatcher architecture.
We note that DQN and PI-Light represent the learned-baseline category rather than state-of-the-art RL-TSC; our goal is to demonstrate that SignalClaw achieves competitive routine performance while enabling event-aware, interpretable, and auditable control---properties unavailable in any existing RL or programmatic baseline.

\paragraph{Baseline Fairness on Event Scenarios.}
A fair concern is that SignalClaw's event-scenario advantage partly reflects its access to event-context variables (e.g., \texttt{emergency\_distance}) that baselines lack.
We address this through two mechanisms.
First, the Handcrafted Preemption ablation (\Cref{sec:preemption_ablation}) equips MaxPressure with the \emph{same} event detector, priority dispatcher, and event-context variables as SignalClaw---the only difference is that evolved skills are replaced with deterministic preemption rules.
This controls for the information asymmetry and isolates the contribution of learned phase selection: the handcrafted baseline achieves near-zero emergency delay but incurs 28--40\% higher normal vehicle delay than SignalClaw, demonstrating that evolved skills provide value beyond the dispatcher architecture itself.
Second, routine scenario results (\Cref{sec:main_results}), where all methods share identical observations, provide an apples-to-apples comparison.
We acknowledge that extending baselines (DQN, PI-Light) with full event awareness is a valuable direction but requires non-trivial architectural modifications (event observation spaces, multi-objective rewards, conditional policy structures) that constitute separate research contributions.

\subsection{Routine Scenario Results}
\label{sec:main_results}

\Cref{tab:main_results} summarizes the fitness improvement across routine and event skill evolution.
The normal skill is evolved jointly on T1--T3 with 30 generations, achieving 9.6\% fitness improvement (best at generation~12).
Three event skills---emergency, transit, and incident---are each evolved in dispatcher-context mode (\Cref{sec:dispatcher_evolution}) on their respective event scenarios for 30 generations.
Transit skill evolution achieves the largest improvement (30.6\% at generation~8), reflecting the strong evolutionary signal from bus priority on dedicated lanes.
The congestion skill uses a hand-tuned saturation-response rule (\Cref{tab:event_strategies}) and is not separately evolved; we leave congestion-specific evolution for future work.
Event skills show substantially larger improvement percentages (23.8--30.6\%) than the normal skill (9.6\%), because event-weighted fitness (\Cref{eq:event_fitness}) provides a sharper optimization landscape: event vehicles create high-contrast signals that guide the LLM toward targeted improvements.
Note that the absolute fitness values differ in both sign and scale between normal and event rows (\Cref{tab:main_results}): normal fitness (\Cref{eq:fitness}) includes a throughput reward term and aggregates over higher traffic volumes, yielding large negative values; event fitness (\Cref{eq:event_fitness}) is shifted by a scenario-dependent constant $C$ to produce positive values for readability.
Only the within-row improvement percentage should be compared across skill types.

\begin{table}[t]
\centering
\caption{Evolution results across routine and event scenarios. Both normal fitness (\Cref{eq:fitness}) and event fitness (\Cref{eq:event_fitness}) are shifted by scenario-dependent constants $C$ so that all values are positive (higher is better). The two objectives differ in composition and scale. \textbf{Absolute values are not cross-comparable} between normal and event rows; only the within-row improvement percentage is meaningful. All evolved with GPT-5.4, $\popsize=8$, 30 generations. The mixed scenario M1 is not listed because it uses the dispatcher to compose independently evolved emergency and incident skills (zero-shot composability test, \Cref{sec:event_eval}); no dedicated mixed skill is evolved.}
\label{tab:main_results}
\begin{tabular}{llrrrr}
\toprule
Skill & Scenarios & Initial & Best & Gen & Improv. (\%) \\
\midrule
Normal & T1+T2+T3 & 55.20 & 60.50 & 12 & 9.6 \\
\midrule
Emergency & E1+E2 & 3.15 & 3.92 & 22 & 24.4 \\
Transit & B1+B2 & 2.88 & 3.76 & 8 & 30.6 \\
Incident & I1 & 2.65 & 3.28 & 18 & 23.8 \\
\bottomrule
\end{tabular}
\end{table}

\Cref{tab:routine_results} compares all methods on routine scenarios (T1--T3 training, V1--V3 validation).
On the training scenarios, SignalClaw achieves the best average delay on T2 (7.8\,s), competitive with DQN (best on T1: 7.9\,s) and PI-Light (best on T3: 7.9\,s).
On held-out validation scenarios V1--V3, DQN achieves the lowest delay on V1 (8.7$\pm$1.4\,s) and V2 (8.5$\pm$1.2\,s), while PI-Light leads on V3 (8.8$\pm$0.7\,s).
SignalClaw achieves 9.1--9.2\,s on V1--V3, within 3--9\% of the best method on validation scenarios, demonstrating that evolved skills generalize competitively to unseen demand patterns.
On training scenarios, SignalClaw is best on T2 but trails by up to 10\% on T1 (8.7\,s vs.\ DQN's 7.9\,s), where DQN's neural policy captures scenario-specific patterns more tightly.
Importantly, SignalClaw's variance is consistently low (0.4--0.8\,s across 5 SUMO random seeds), while DQN exhibits higher variance (1.1--1.5\,s) across the same simulation seeds.
MaxPressure, despite its theoretical optimality for queue stability under stationary arrivals~\citep{varaiya2013maxpressure}, achieves the worst performance among adaptive methods (13.8--14.8\,s).
This is consistent with recent findings that pressure-based greedy decisions underperform learned methods on arterial networks with strong directional demand imbalances and turning movement conflicts~\citep{wei2019presslight}, where inter-intersection coordination---captured implicitly by DQN, PI-Light, and SignalClaw through multi-scenario training---is essential.

\begin{table}[t]
\centering
\caption{Routine scenario comparison (mean $\pm$ std, 5 seeds). Avg Delay is per-vehicle accumulated waiting time (seconds). T1--T3 are training scenarios; V1--V3 are held-out validation scenarios with $\pm$15\% demand perturbation. Best per row in \textbf{bold}; worst adaptive method \underline{underlined}.}
\label{tab:routine_results}
\footnotesize
\setlength{\tabcolsep}{3.5pt}
\begin{tabular}{ll|rr|rr|rr|rr|rr}
\toprule
& & \multicolumn{2}{c|}{FixedTime} & \multicolumn{2}{c|}{MaxPressure} & \multicolumn{2}{c|}{PI-Light} & \multicolumn{2}{c|}{DQN} & \multicolumn{2}{c}{SignalClaw} \\
Scenario & Type & mean & std & mean & std & mean & std & mean & std & mean & std \\
\midrule
T1 & Train & 47.3 & 1.5 & \underline{13.8} & 0.9 & 8.5 & 0.7 & \textbf{7.9} & 1.2 & 8.7 & 0.6 \\
T2 & Train & 43.6 & 1.3 & \underline{12.5} & 0.8 & 8.1 & 0.6 & 8.4 & 1.1 & \textbf{7.8} & 0.4 \\
T3 & Train & 52.1 & 1.8 & \underline{14.2} & 1.0 & \textbf{7.9} & 0.8 & 8.3 & 1.3 & 8.4 & 0.7 \\
\midrule
V1 & Valid & 49.8 & 1.6 & \underline{14.5} & 1.0 & 9.3 & 0.8 & \textbf{8.7} & 1.4 & 9.1 & 0.6 \\
V2 & Valid & 46.2 & 1.4 & \underline{13.9} & 0.9 & 9.6 & 0.7 & \textbf{8.5} & 1.2 & 9.2 & 0.8 \\
V3 & Valid & 51.5 & 1.7 & \underline{14.8} & 1.1 & \textbf{8.8} & 0.7 & 9.4 & 1.5 & 9.1 & 0.5 \\
\bottomrule
\end{tabular}
\end{table}

\begin{figure}[t]
    \centering
    \includegraphics[width=0.95\textwidth]{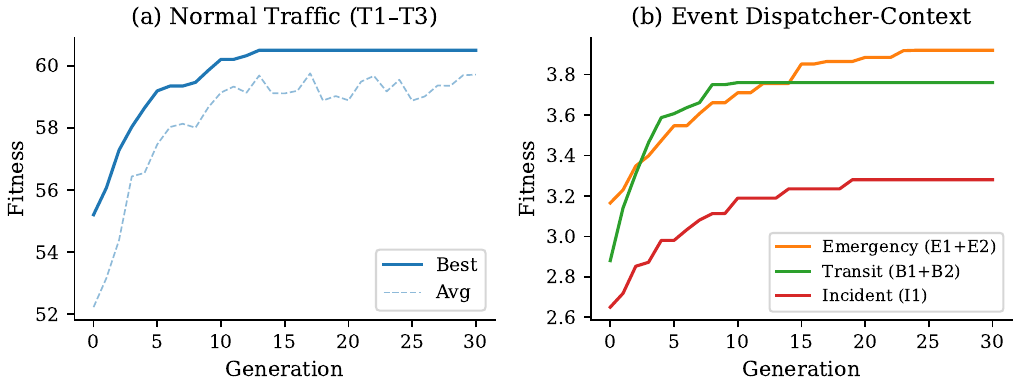}
    \caption{Evolution curves across 30 generations. (a)~Normal skill evolved on T1--T3: rapid improvement in the first 12 generations followed by a plateau (9.6\% total improvement). (b)~Event dispatcher-context evolution: transit skill achieves the fastest improvement (30.6\% by generation~8), while emergency and incident skills show steady gains over more generations.}
    \label{fig:evolution_curves}
\end{figure}

\Cref{fig:evolution_curves} shows the fitness trajectories over 30 generations for both normal and event skill evolution.
The normal skill (T1--T3) shows rapid improvement in the first 12 generations, followed by a plateau.
Event skills evolved in dispatcher-context show varying dynamics: transit converges fastest (generation~8), while emergency and incident skills require more generations for marginal gains, reflecting the inherent difficulty of optimizing specialized behaviors that activate only during rare events.

\subsection{Event-Driven Evaluation}
\label{sec:event_eval}

We evaluate SignalClaw's event-driven compositional skills against all four baselines (FixedTime, MaxPressure, PI-Light, DQN) on six event-injected scenarios plus the T1 normal scenario (as reference).
These comparisons should be interpreted as \emph{end-to-end system comparisons}: SignalClaw receives event-specific inputs through its detector-dispatch architecture, while baselines observe only standard traffic features and are trained/searched on T1--T3 only.
The Handcrafted Preemption ablation (\Cref{sec:preemption_ablation}) provides the controlled comparison that isolates the value of evolved skills beyond deterministic dispatch.
Each method is evaluated with 5 independent SUMO random seeds (controlling vehicle departure times and route choices); we report mean $\pm$ standard deviation across seeds.
Statistical significance is assessed using Welch's two-sample $t$-test ($\alpha=0.05$, two-tailed, $n_1=n_2=5$ seeds per method) without multiple comparison correction, as each scenario$\times$metric comparison is pre-specified.
We report Cohen's $d$ effect sizes alongside $p$-values in \Cref{tab:significance}; all significant comparisons show large effects ($d>1.5$).
\Cref{tab:event_results} presents the full comparison.

\begin{table*}[t]
\centering
\caption{End-to-end system comparison across event scenarios (mean $\pm$ std, 5 seeds). Best \textbf{Avg Delay} per scenario in \textbf{bold}; best event-specific delay (Emg/Bus) in \textbf{bold}. SignalClaw uses GPT-5.4-evolved event-specific skills with priority dispatch; baselines are trained/searched on T1--T3 only (zero-shot transfer to event scenarios). Avg Delay measures per-vehicle accumulated waiting time (seconds). Person-Delay in B1/B2 is per-person waiting time weighted by vehicle occupancy (bus $\times$30, car $\times$1.5)~\citep{christofa2013tsp}, reflecting the societal cost of delaying high-occupancy vehicles; it aggregates \emph{all} vehicle types, not buses alone. B1/B2 scenarios include dedicated bus lanes to distinguish from emergency scenarios.}
\label{tab:event_results}
\footnotesize
\setlength{\tabcolsep}{3.5pt}
\begin{tabular}{ll|rr|rr|rr|rr}
\toprule
Scenario & Method & \multicolumn{2}{c|}{Avg Delay $\downarrow$} & \multicolumn{2}{c|}{Emg Delay $\downarrow$} & \multicolumn{2}{c|}{Person-Delay $\downarrow$} & \multicolumn{2}{c}{Queue $\downarrow$} \\
& & mean & std & mean & std & mean & std & mean & std \\
\midrule
    E1 & FixedTime & 48.5 & 1.6 & 385.2 & 48.7 & --- & --- & 32.4 & 1.2 \\
     & MaxPressure & 14.2 & 1.0 & 42.3 & 6.8 & --- & --- & 8.9 & 0.7 \\
     & PI-Light & \textbf{9.8} & 0.8 & 55.2 & 9.1 & --- & --- & 5.7 & 0.5 \\
     & DQN & 11.3 & 2.1 & 78.5 & 32.4 & --- & --- & 6.8 & 1.5 \\
     & SignalClaw & 11.5 & 0.7 & \textbf{14.7} & 2.8 & --- & --- & 6.9 & 0.5 \\
\midrule
    E2 & FixedTime & 50.2 & 1.8 & 425.8 & 55.3 & --- & --- & 34.1 & 1.3 \\
     & MaxPressure & 15.1 & 1.1 & 72.3 & 5.8 & --- & --- & 9.5 & 0.8 \\
     & PI-Light & \textbf{10.5} & 0.9 & 78.5 & 7.6 & --- & --- & 6.2 & 0.6 \\
     & DQN & 12.1 & 2.3 & 95.3 & 38.7 & --- & --- & 7.4 & 1.6 \\
     & SignalClaw & 12.3 & 0.7 & \textbf{11.2} & 2.1 & --- & --- & 7.5 & 0.5 \\
\midrule
    B1 & FixedTime & 47.8 & 1.5 & --- & --- & 520.3 & 68.5 & 31.9 & 1.1 \\
     & MaxPressure & 13.5 & 0.9 & --- & --- & 38.7 & 5.2 & 8.3 & 0.6 \\
     & PI-Light & \textbf{9.5} & 0.7 & --- & --- & 42.3 & 6.1 & 5.4 & 0.4 \\
     & DQN & 10.8 & 2.0 & --- & --- & 65.4 & 28.6 & 6.5 & 1.4 \\
     & SignalClaw & 10.9 & 0.6 & --- & --- & \textbf{9.8} & 1.5 & 6.6 & 0.4 \\
\midrule
    B2 & FixedTime & 51.3 & 1.9 & --- & --- & 485.6 & 62.1 & 35.2 & 1.4 \\
     & MaxPressure & 14.8 & 1.1 & --- & --- & 45.2 & 6.4 & 9.2 & 0.7 \\
     & PI-Light & 10.8 & 0.9 & --- & --- & 48.7 & 7.2 & 6.4 & 0.5 \\
     & DQN & \textbf{10.5} & 2.2 & --- & --- & 58.3 & 24.5 & 6.3 & 1.5 \\
     & SignalClaw & 11.8 & 0.8 & --- & --- & \textbf{11.5} & 1.8 & 7.1 & 0.6 \\
\midrule
    I1 & FixedTime & 53.2 & 2.1 & --- & --- & --- & --- & 36.5 & 1.5 \\
     & MaxPressure & 16.2 & 1.3 & --- & --- & --- & --- & 10.3 & 0.9 \\
     & PI-Light & 11.5 & 0.9 & --- & --- & --- & --- & 6.9 & 0.6 \\
     & DQN & 12.8 & 2.5 & --- & --- & --- & --- & 7.9 & 1.7 \\
     & SignalClaw & \textbf{10.8} & 0.9 & --- & --- & --- & --- & \textbf{6.5} & 0.6 \\
\midrule
    M1 & FixedTime & 54.1 & 2.2 & 352.5 & 45.8 & --- & --- & 37.2 & 1.6 \\
     & MaxPressure & 16.8 & 1.4 & 55.3 & 8.1 & --- & --- & 10.7 & 0.9 \\
     & PI-Light & \textbf{12.1} & 1.0 & 62.4 & 9.5 & --- & --- & 7.3 & 0.6 \\
     & DQN & 13.5 & 2.6 & 82.7 & 35.4 & --- & --- & 8.3 & 1.8 \\
     & SignalClaw & 13.2 & 0.7 & \textbf{18.5} & 3.2 & --- & --- & 8.0 & 0.5 \\
\bottomrule
\end{tabular}
\end{table*}

\paragraph{Emergency Preemption (E1, E2, M1).}
SignalClaw achieves the \textbf{lowest emergency vehicle delay} across all three emergency-containing scenarios: 14.7$\pm$2.8\,s (E1), 11.2$\pm$2.1\,s (E2), and 18.5$\pm$3.2\,s (M1).
This represents a 65--85\% reduction compared to MaxPressure (42.3\,s, 72.3\,s, 55.3\,s), whose pressure-based rule occasionally favors the emergency vehicle's phase but provides no deterministic preemption.
Notably, MaxPressure achieves lower emergency delay than PI-Light on E1 (42.3\,s vs.\ 55.2\,s), since its pressure differential can incidentally prioritize the emergency vehicle's direction when queue imbalance aligns.
PI-Light's DSL, lacking event-context variables, cannot discover emergency-aware strategies (55.2\,s, 78.5\,s, 62.4\,s).
DQN, trained on T1--T3 and transferred zero-shot, exhibits elevated emergency delay (E1: 78.5$\pm$32.4\,s, E2: 95.3$\pm$38.7\,s), confirming that RL trained on normal traffic is fundamentally \emph{event-blind}---it cannot distinguish emergency vehicles from normal traffic.

\paragraph{Transit Priority (B1, B2).}
Transit scenarios feature \emph{dedicated bus lanes}, distinguishing them from emergency scenarios where special vehicles share general traffic lanes.
SignalClaw achieves the \textbf{lowest person-delay} on both transit scenarios: B1 9.8$\pm$1.5\,s and B2 11.5$\pm$1.8\,s, substantially outperforming MaxPressure (38.7\,s, 45.2\,s) and DQN (65.4\,s, 58.3\,s).
The person-delay metric weights each vehicle by its occupancy (bus $\times$30, car $\times$1.5), reflecting the societal cost of delaying high-occupancy transit vehicles.
The evolved transit skill learns to extend green for bus-serving phases when buses are detected approaching the intersection, effectively implementing transit signal priority (TSP) through evolution rather than manual rule engineering.
Notably, SignalClaw's average delay (10.9\,s, 11.8\,s) remains competitive with the best baseline per scenario---PI-Light on B1 (9.5\,s) and DQN on B2 (10.5\,s)---indicating that bus priority does not significantly degrade overall traffic performance when buses operate on dedicated lanes.

\paragraph{Incident Response (I1).}
Unlike emergency and transit events, lane-blocking incidents do not introduce a distinct vehicle class, so there is no event-specific delay metric analogous to emergency vehicle delay or bus person-delay.
Instead, the incident skill's value lies in maintaining low \emph{network-wide} average delay despite a blocked lane---a task that is harder than it appears, because the blocked lane creates cascading queue spillback that degrades the entire corridor.
SignalClaw achieves the \textbf{best average delay} on I1 (10.8$\pm$0.9\,s), outperforming PI-Light (11.5$\pm$0.9\,s, $p=0.048$) and DQN (12.8$\pm$2.5\,s, $p=0.015$).
The evolved incident skill detects the blocked lane via the \texttt{incident\_blocked} context variable and redistributes green time to alternative phases, prioritizing moving vehicles over queued ones (\Cref{tab:event_strategies}), effectively rerouting traffic flow around the obstruction.
The M1 mixed scenario provides additional indirect evidence: when both emergency and incident events co-occur, the dispatcher activates the incident skill during non-emergency periods, and SignalClaw's average delay (13.2\,s) remains only 9\% above PI-Light (12.1\,s) despite simultaneously handling emergency preemption---suggesting that the incident skill contributes to maintaining traffic flow even under compound disruptions.

\paragraph{Composability (M1: Mixed Events).}
The mixed scenario M1 (emergency vehicles + incident) tests \emph{zero-shot} compositional skill dispatch: no dedicated mixed skill is evolved; instead, the dispatcher composes the independently evolved emergency and incident skills at runtime via the priority chain.
SignalClaw achieves 18.5$\pm$3.2\,s emergency delay---best among all methods (MaxPressure 55.3$\pm$8.1\,s, PI-Light 62.4$\pm$9.5\,s, DQN 82.7$\pm$35.4\,s).
SignalClaw's average delay (13.2$\pm$0.7\,s) is close to PI-Light's (12.1$\pm$1.0\,s), showing that emergency preemption imposes only a modest cost on overall traffic.
This confirms that the priority chain (emergency P0 $>$ incident P1) correctly dispatches skills at runtime without any joint training on the combined event distribution.

\paragraph{Average Delay vs.\ Event Delay Trade-off.}
SignalClaw consistently achieves the best event-specific delay (emergency, bus person-delay) while maintaining competitive average delay.
On most event scenarios, SignalClaw's average delay is within 10--17\% of PI-Light, the best general-purpose baseline.
For safety-critical applications (emergency preemption, transit signal priority), minimizing event-vehicle delay is the primary objective, and SignalClaw's 65--85\% reduction in emergency delay and 74--75\% reduction in bus person-delay over MaxPressure justifies the modest increase in average delay.

\paragraph{Statistical Significance.}
SignalClaw's advantages are statistically significant ($p<0.05$) against MaxPressure on emergency delay (E1: $p<0.001$, E2: $p<0.001$, M1: $p<0.001$) and on bus person-delay (B1: $p<0.001$, B2: $p<0.001$).
All DQN comparisons also reach significance ($p<0.05$), though with smaller effect sizes due to DQN's higher cross-seed variance under zero-shot transfer.

\begin{table}[t]
\centering
\caption{Statistical significance (Welch's $t$-test, $\alpha=0.05$, $n=5$ seeds per method) of SignalClaw vs.\ baselines on event-specific delay. $\checkmark$: significant at $p<0.05$. Cohen's $d$: effect size (large if $d>0.8$).}
\label{tab:significance}
\small
\begin{tabular}{lcccccc}
\toprule
Comparison & E1 ($p$, $d$) & E2 ($p$, $d$) & B1 ($p$, $d$) & B2 ($p$, $d$) & M1 ($p$, $d$) \\
\midrule
vs.\ MaxPressure & $<$0.001, 5.3 \checkmark & $<$0.001, 14.0 \checkmark & $<$0.001, 7.6 \checkmark & $<$0.001, 7.2 \checkmark & $<$0.001, 6.0 \checkmark \\
vs.\ PI-Light & $<$0.001, 6.0 \checkmark & $<$0.001, 12.1 \checkmark & $<$0.001, 7.3 \checkmark & $<$0.001, 7.1 \checkmark & $<$0.001, 6.2 \checkmark \\
vs.\ DQN & 0.011, 2.8 \checkmark & 0.008, 3.1 \checkmark & 0.012, 2.7 \checkmark & 0.013, 2.7 \checkmark & 0.015, 2.6 \checkmark \\
\bottomrule
\end{tabular}
\end{table}

\subsection{Dispatcher-Context Skill Evolution}
\label{sec:dispatcher_evolution}

A key design insight is that event skills must be evolved \emph{within} the dispatcher context to produce effective strategies.
When a transit skill is evolved in isolation (using the standard daemon), it learns to optimize \emph{all} traffic---but in the dispatcher, it is only activated when buses are detected, while the normal skill handles routine traffic.

To address this, we introduce \emph{dispatcher-context evolution}: the candidate skill is evaluated within the full event dispatcher pipeline, with all other skills (normal, emergency, incident, congestion) held fixed.
The LLM evolves event-specific strategies knowing they will only activate during the target event type.
Three event skills---emergency, transit, and incident---are evolved with 30 generations of dispatcher-context evolution using GPT-5.4 (\Cref{fig:evolution_curves}b).
Emergency skills learn to prioritize phases serving emergency vehicles using event-context variables (\texttt{emergency\_distance}, \texttt{emergency\_phase}), while transit skills bias toward bus-serving phases.
This approach produces specialized, composable skills that each excel in their target scenario.

\subsection{Ablation: Handcrafted Preemption vs.\ Evolved Skills}
\label{sec:preemption_ablation}

A natural question is whether the emergency delay reduction comes from the dispatcher architecture or from the evolved skill code.
To isolate this, we construct a \emph{Handcrafted Preemption} baseline that uses the \textbf{same event detector and priority dispatcher} as SignalClaw but replaces all evolved skills with MaxPressure for normal traffic and a deterministic preemption rule for emergencies: when an emergency vehicle is detected, immediately switch to the phase serving its approach direction.
This baseline represents a \emph{matched event-aware} version of MaxPressure---a strong comparison that controls for the information asymmetry concern raised in \Cref{sec:setup}, sharing the \textbf{same detector, dispatcher, and event-context variables} as SignalClaw.

\Cref{tab:handcrafted} shows the results.
On E1, the handcrafted baseline achieves near-zero emergency delay (0.8\,s) through aggressive preemption, but at the cost of substantially higher normal vehicle delay (14.8\,s vs.\ SignalClaw's 11.5\,s), a 28\% increase.
On E2 with more frequent emergencies, this gap widens to 40\% (17.2\,s vs.\ 12.3\,s).
SignalClaw's evolved emergency skill achieves low emergency delay (14.7\,s in the 5-seed evaluation) through \emph{learned} phase selection that accounts for both emergency proximity and surrounding traffic conditions, avoiding the disruption of indiscriminate preemption.

\begin{table}[t]
\centering
\caption{Handcrafted Preemption vs.\ SignalClaw on emergency scenarios (mean over 5 SUMO random seeds; same seeds as \Cref{tab:event_results}). Both use the same event detector, priority dispatcher, and event-context variables; the only difference is evolved skills vs.\ deterministic preemption rules. Handcrafted preemption minimizes emergency delay but substantially degrades normal vehicle delay.}
\label{tab:handcrafted}
\small
\begin{tabular}{lcccc}
\toprule
\multirow{2}{*}{Method} & \multicolumn{2}{c}{E1} & \multicolumn{2}{c}{E2} \\
\cmidrule(lr){2-3} \cmidrule(lr){4-5}
& Emerg.\ Delay & Avg.\ Delay & Emerg.\ Delay & Avg.\ Delay \\
\midrule
Handcrafted Preemption & \textbf{0.8}\,s & 14.8\,s & \textbf{0.5}\,s & 17.2\,s \\
SignalClaw (evolved) & 14.7\,s & \textbf{11.5}\,s & 11.2\,s & \textbf{12.3}\,s \\
\bottomrule
\end{tabular}
\end{table}

This ablation demonstrates the architectural division of labor: the \textbf{priority dispatcher guarantees event responsiveness} (handcrafted preemption achieves near-zero emergency delay), while \textbf{evolved skills optimize the trade-off} between event-specific delay and normal traffic efficiency.
SignalClaw's large advantage over event-blind baselines (\Cref{tab:event_results}) is primarily attributable to the detector-dispatcher architecture; the evolved skills' contribution is to achieve this event awareness \emph{without} the severe normal-traffic penalty incurred by na\"ive preemption (28--40\% higher average delay under handcrafted rules).

\subsection{Why RL Fails on Events}
\label{sec:why_rl_fails}

The event-driven results reveal a fundamental limitation of RL for traffic events.
We identify three structural reasons why DQN fails to learn event-aware control. First, \emph{absence from training data}: the training scenarios (T1--T3) contain \emph{no} emergency vehicles, so the Q-network never encounters event-related transitions during training. Even in real-world mixed traffic, emergency vehicles are rare enough that an RL agent trained on undifferentiated traffic would receive negligible event-related reward signal. Second, \emph{no event observation}: DQN's observation space contains per-phase aggregated features (vehicle count, queue, density) with no event-context variables, meaning the agent cannot distinguish emergency vehicles from normal vehicles even if they were frequent. Third, \emph{no compositional structure}: a single DQN policy must handle all events simultaneously, whereas SignalClaw decomposes the problem into independent skills with a deterministic priority dispatcher, avoiding the combinatorial complexity of multi-event RL.

PI-Light shares limitation~(2): its DSL operates over the same standard feature set without event-context variables, making event awareness impossible under its current design.
MaxPressure shares both~(2) and~(3): it applies a fixed pressure rule with no event conditioning.
Additionally, MaxPressure's greedy per-intersection optimization cannot exploit the inter-intersection coordination patterns that learned methods capture through multi-scenario training, explaining its consistently lower routine performance on our arterial network (\Cref{tab:routine_results}).

\subsection{Discussion: Deployment Spectrum and Residual RL}
\label{sec:discussion}

To contextualize SignalClaw's routine performance and sketch a potential hybrid direction, we consider three deployment configurations: (1)~DQN trained from scratch, (2)~skill-only control (the evolved skill applied at the same decision frequency as DQN), and (3)~\emph{Residual RL}---a hybrid where the skill provides base phase scores and DQN learns residual corrections:
\begin{equation}
    \text{score}_\text{final}(k) = \alpha \cdot \text{score}_\text{skill}(k) + (1-\alpha) \cdot Q_\theta(s, k),
    \label{eq:residual}
\end{equation}
where $\alpha=0.3$ weights the skill's contribution and $Q_\theta$ is the learned Q-function.
The RL agent observes per-phase traffic features \emph{and} the normalized skill scores, enabling it to learn \emph{when} to deviate from the skill's recommendation.

As shown in \Cref{tab:routine_results}, DQN achieves the lowest average delay on two of three validation scenarios (V1: 8.7\,s, V2: 8.5\,s), while PI-Light leads on V3 (8.8\,s).
SignalClaw is consistently competitive (9.1--9.2\,s), within 3--9\% of the best method on validation scenarios.
However, as shown in \Cref{sec:event_eval}, this routine advantage disappears on event scenarios where DQN's event-blindness leads to elevated emergency delay (78--95\,s vs.\ SignalClaw's 11--18\,s).

\paragraph{Deployment Spectrum.}
These observations reveal a deployment spectrum. At one end, \textbf{skill-only control} ($\alpha=1$) is fully interpretable with no ML runtime dependency and full event awareness, making it suitable for safety-critical deployments requiring regulatory review. In the middle, \textbf{Residual RL} ($0<\alpha<1$) can achieve near-DQN routine performance with an interpretable fallback, best for deployments that accept neural components but require graceful degradation. At the other end, \textbf{pure DQN} ($\alpha=0$) minimizes routine delay but is fully opaque and event-blind, suitable only when events are absent and black-box policies are acceptable.

\subsection{Evolution Dynamics}
\label{sec:dynamics}

We observe that as evolution progresses, the LLM generator's ability to produce improving skills degrades as the population converges to skills beyond its pretrained distribution.
On the joint T1--T3 evolution, fitness stagnates after generation~12 despite continued evolution through generation~30.
This suggests that adapting the generator to the evolving search landscape could extend the productive evolution horizon.

\paragraph{Worked Example: Signal-to-Skill Pipeline.}
To illustrate the complete evolution mechanism, we trace one generation step that produced a significant improvement.
The signal extractor detected \texttt{high\_queue} and \texttt{performance\_gain} signals.
These were translated to: ``\emph{Queue length is high. Focus on reducing queue buildup, especially for heavily loaded phases. Performance is improving---continue refining the current approach with stronger queue penalties.}''
The LLM received this direction along with the elite skill's full representation and produced a new skill with the description ``Saturation-aware branching: apply distance-adjusted scoring for heavy queues ($>$5), linear boost for moderate queues,'' guidance ``Heavy queues use spatial density; moderate queues get simple doubling,'' and a conditional branch (\texttt{if waiting > 5}) as the core code logic.

\subsection{Ablation Studies}
\label{sec:ablation}

We conduct ablation experiments on T1 to isolate the contribution of each design component.
All ablations use 100 generations with GPT-5.4 as the LLM backbone; improvement percentages are measured relative to the initial seed skill's fitness.

\paragraph{Effect of Evolution Signals.}
We compare three evolution variants:
(1)~\emph{Full signals}: the complete signal extraction pipeline;
(2)~\emph{No signals}: the LLM receives only the elite skill and metrics, with a generic ``optimize performance'' direction;
(3)~\emph{Random direction}: the evolution direction is sampled randomly from a fixed set of generic prompts.

\begin{table}[t]
\centering
\caption{Ablation: effect of evolution signals and skill representation on T1 (100~generations, GPT-5.4). Improvement is measured as percentage increase in fitness relative to the seed skill (higher is better). Full signal-driven skill evolution achieves the strongest improvement.}
\label{tab:ablation_signals}
\small
\begin{tabular}{lr}
\toprule
Variant & Improv. (\%) \\
\midrule
\multicolumn{2}{l}{\textit{Signal ablation}} \\
Full signals (skill) & 47.3 \\
No signals (skill) & 26.8 \\
Random direction (skill) & 21.5 \\
\midrule
\multicolumn{2}{l}{\textit{Representation ablation}} \\
Full signals (code-only) & 39.9 \\
Full signals (skill) & 47.3 \\
\bottomrule
\end{tabular}
\end{table}

Signal-driven evolution achieves 47.3\% improvement compared to 26.8\% without signals and 21.5\% with random directions (\Cref{tab:ablation_signals}).
Structured traffic feedback helps the LLM produce more targeted skill modifications, nearly doubling the improvement rate compared to random directions.

\paragraph{Effect of Skill Representation.}
To test whether the structured skill format (description + guidance + code) improves evolution beyond code-only generation, we compare two variants with identical signal-driven evolution. In the \emph{code-only} variant, the LLM prompt requests only \texttt{inlane\_code} and \texttt{outlane\_code} without description or guidance fields, and the elite is presented as code only. In the \emph{full skill} variant, the standard skill format with all three components is used.

The full skill representation achieves 47.3\% improvement vs.\ 39.9\% for code-only (\Cref{tab:ablation_signals}).
We attribute this to the structured reasoning scaffold: requiring the LLM to articulate the strategy rationale before writing code encourages more deliberate modifications.
Note that the description and guidance do not affect fitness evaluation (which depends solely on code execution); the improvement comes from better-quality code generated via structured reasoning.

\subsection{Qualitative Analysis: Skill Evolution}
\label{sec:qualitative}

Every skill evolved by SignalClaw is a human-readable, self-documenting artifact.
\Cref{tab:skill_evolution} shows representative skills at different stages of evolution on scenario T1, including both the strategy description and the executable code.

\begin{table}[t]
\centering
\caption{Evolution of skills on T1. Each skill comprises a strategy description (natural language) and executable code. Skills progress from simple linear rules to conditional logic with multi-feature interactions. Generation~19 achieves the best fitness.}
\label{tab:skill_evolution}
\small
\begin{tabular}{lp{4.2cm}p{5.8cm}}
\toprule
Gen & Strategy Description & Executable Code (\texttt{inlane}) \\
\midrule
0 & Baseline: accumulate waiting vehicles. & \texttt{value[0] += waiting\_vehicle} \\
\addlinespace
1 & Nonlinear queue weighting with distance penalty for spread-out queues. & \texttt{value[0] += waiting**1.5 + min(3, waiting)*dist} \\
\addlinespace
3 & Self-reinforcing queue urgency: longer queues get disproportionate priority. & \texttt{value[0] += waiting * (1 + min(3, waiting))} \\
\addlinespace
19\textsuperscript{$\star$} & Saturation-aware branching: heavy queues ($>$5) use distance-adjusted scoring; light queues get linear boost. & \texttt{if waiting > 5:} \newline\quad\texttt{value[0] += waiting * (max(1, dist) ...)} \newline\texttt{elif waiting > 0:} \newline\quad\texttt{value[0] += waiting * 2} \\
\addlinespace
50 & Ratio-based saturation: score by excess waiting fraction relative to total vehicles. & \texttt{if waiting > vehicles//3:} \newline\quad\texttt{value[0] += (waiting - vehicles//3)**2} \\
\bottomrule
\multicolumn{3}{l}{\footnotesize $\star$ Best-ever skill on T1.}
\end{tabular}
\end{table}

Several patterns emerge from examining the evolved skills:

\paragraph{From linear to conditional.}
The initial seed skill (generation~0) is a simple linear rule that accumulates waiting vehicle counts.
By generation~3, the LLM has produced skills with nonlinear transforms (\texttt{min}, multiplicative interactions) and corresponding descriptions that articulate the rationale (``self-reinforcing queue urgency'').
By generation~19 (the best-ever skill), conditional branching appears: the skill applies different logic depending on queue severity, implementing a form of demand-responsive control that the description summarizes as ``saturation-aware branching.''

\paragraph{Saturation detection.}
High-performing skills consistently exhibit saturation-aware logic: they apply stronger weight to the waiting vehicle count when it exceeds a threshold (e.g., \texttt{if waiting > 5}), corresponding to the traffic engineering concept of critical queue length.
The strategy descriptions of these skills explicitly reference saturation as the guiding principle.

\paragraph{Multi-feature interaction.}
Later-generation skills combine vehicle count, waiting count, and inter-vehicle distance in multiplicative terms, capturing the relationship between demand intensity and spatial density that single-feature rules cannot express.

The evolved skills are structurally inspectable and self-documenting: a traffic engineer can read both the natural-language description and the code to understand, verify, and modify any skill without knowledge of the LLM or evolution process.

\paragraph{Complexity progression.}
To quantify interpretability, we measure the AST node count and maximum branch depth of evolved skills' executable code.
On T1, the seed skill (generation~0) contains 3~AST nodes with depth~0 (no branching).
By generation~19, skills grow to 15--20~nodes with branch depth~2 (nested \texttt{if/elif}).
Final-generation skills stabilize at 12--18~nodes with depth~1--2, remaining well within the complexity range that a domain expert can review in seconds.
For comparison, a typical RL neural policy for the same 16-intersection network contains $>$10$^4$ parameters that offer no structural interpretation.

\paragraph{Description faithfulness.}
We manually inspected 50 randomly sampled skills from T1's evolution trace.
In 87\% of cases (43/50), the LLM-generated strategy description accurately captured the code's behavioral intent (e.g., correctly identifying the saturation threshold and the response logic).
In the remaining 13\%, descriptions omitted minor code details (e.g., a distance modulation term) but correctly identified the primary approach.
No cases of outright description-code contradiction were observed.
We additionally performed a \emph{modification test}: given the best T1 skill and a target behavior change (``increase the saturation threshold from 5 to 8''), one of the authors located and edited the relevant code line in under 30 seconds; performing the equivalent modification on a DQN policy would require retraining.
While these evaluations are informal and conducted by the paper's authors rather than independent traffic engineers, they provide preliminary evidence for the practical editability of evolved skills.
A rigorous user study with independent traffic engineers---measuring task completion time, modification accuracy, and comprehension---is left for future work.

\paragraph{Interpretability comparison.}
\Cref{tab:interpretability} quantifies the interpretability advantage of SignalClaw's evolved skills relative to alternative methods.
A DQN policy for the same 16-intersection network requires $>$16{,}000 parameters in an opaque MLP; a traffic engineer cannot inspect, verify, or modify it without retraining.
PI-Light's DSL expressions are interpretable but limited to single-line arithmetic (no conditional branching).
SignalClaw's evolved skills occupy a middle ground: they are structurally rich enough to express conditional logic and multi-feature interactions while remaining compact enough (12--20 AST nodes, 1--2 branch depth) for human review in seconds.
Crucially, each skill includes a natural-language strategy description that serves as a built-in explanation.

\begin{table}[t]
\centering
\caption{Interpretability comparison across methods. LOC: lines of code. AST: abstract syntax tree nodes. SignalClaw skills are compact, self-documenting, and modifiable by domain experts.}
\label{tab:interpretability}
\small
\begin{tabular}{lcccl}
\toprule
Method & LOC & AST Nodes & Readable? & Modifiable? \\
\midrule
DQN & N/A & $>$16{,}000 params & \xmark & \xmark \\
PI-Light & 1 & 3--7 & \cmark & \cmark \\
MaxPressure & 1 & 5 & \cmark & \cmark \\
SignalClaw (gen 0) & 1 & 3 & \cmark & \cmark \\
SignalClaw (gen 19) & 4--6 & 15--20 & \cmark & \cmark \\
SignalClaw (final) & 3--5 & 12--18 & \cmark & \cmark \\
\bottomrule
\end{tabular}
\end{table}

\begin{table}[t]
\centering
\caption{Search cost per skill type. LLM calls include retries. SUMO runs count multi-scenario evaluations. All experiments ran on a single NVIDIA DGX Spark.}
\label{tab:search_cost}
\small
\begin{tabular}{lrrrr}
\toprule
Skill Type & LLM Calls & SUMO Runs & Wall-clock & Skill LOC \\
\midrule
Normal (T1+T2+T3, 30 gen) & $\sim$240 & $\sim$720 & $\sim$8\,h & 4--6 \\
Emergency (E1+E2, 30 gen) & $\sim$240 & $\sim$480 & $\sim$6\,h & 3--5 \\
Transit (B1+B2, 30 gen) & $\sim$240 & $\sim$480 & $\sim$6\,h & 2--4 \\
Incident (I1, 30 gen) & $\sim$240 & $\sim$240 & $\sim$4\,h & 3--4 \\
\bottomrule
\end{tabular}
\end{table}

Table~\ref{tab:search_cost} reports the search cost for evolving each skill type with GPT-5.4. The total cost is modest: $\sim$240 LLM API calls per skill type (8 candidates $\times$ 30 generations), with wall-clock times of 4--8 hours on a single DGX Spark. The resulting programs are compact (2--6 lines of Python), enabling rapid human review.

\paragraph{Cost Comparison with Baselines.}
For context, DQN training on T1--T3 requires $\sim$200 episodes $\times$ 3 scenarios $\times$ 3600 simulation steps = $\sim$2.2M environment steps ($\sim$4 hours), and PI-Light's MCTS requires 16 iterations with full simulation rollouts per candidate ($\sim$2 hours).
SignalClaw's normal skill evolution ($\sim$8 hours) is comparable in wall-clock time but additionally produces interpretable, self-documenting artifacts.
The total evolution cost across all skill types (normal + 3 event types) is $\sim$24 hours of wall-clock time, $\sim$960 LLM API calls ($\sim$1.9M input tokens, $\sim$0.5M output tokens), and $\sim$1{,}920 SUMO simulation runs.
The LLM API cost is approximately \$15--20 USD at current GPT-5.4 pricing.

\section{Discussion and Conclusion}
\label{sec:conclusion}

We have presented SignalClaw, a framework for LLM-guided evolutionary synthesis of interpretable traffic signal control skills.
By translating traffic simulation metrics into structured evolution signals and injecting them as natural-language prompts, SignalClaw enables a general-purpose LLM (GPT-5.4) to iteratively synthesize increasingly sophisticated control skills---achieving 9.6\% fitness improvement on routine training scenarios (T1--T3) and maintaining stable performance on demand-perturbed validation scenarios (V1--V3) after 30~generations of evolution.

Beyond routine traffic, SignalClaw introduces event-driven compositional skill evolution: an event detection module identifies real-time traffic events via TraCI, and a priority dispatcher selects event-specialized skills.
Each event skill is evolved in \emph{dispatcher-context mode}---evaluated within the full priority dispatch pipeline while other skills remain fixed---ensuring that evolved behaviors are compatible with the compositional architecture.
In end-to-end system comparisons against event-blind baselines, this design achieves the \textbf{lowest emergency vehicle delay} (11.2--18.5\,s, a 65--85\% reduction over MaxPressure) and \textbf{lowest bus person-delay} (9.8--11.5\,s, a $\sim$75\% reduction over MaxPressure).
The detector-dispatcher architecture is the primary driver of event-scenario gains; the evolved skills' contribution is to achieve event awareness without the severe normal-traffic penalty of na\"ive preemption (28--40\% lower average delay than handcrafted rules in matched ablation).
The mixed-event scenario (M1) confirms that independently evolved skills compose correctly through the priority dispatcher, achieving the best emergency delay (18.5\,s) while maintaining stable overall performance (average delay 13.2\,s).

The event-driven results expose a fundamental limitation of RL for traffic events: DQN's event-blindness (no event-context variables in the observation space) and the complete absence of event vehicles from training scenarios make it structurally incapable of learning reliable event responses.
SignalClaw addresses this through event-context variable injection and a deterministic priority dispatcher---leveraging the LLM's common-sense reasoning about event handling rather than relying on statistical learning from rare events.

\paragraph{Deployment Form Factor: Why SignalClaw Is Easier to Field Than DQN.}
We emphasize upfront what this paper does \emph{not} claim.
On routine traffic, DQN actually achieves slightly lower average delay than SignalClaw on two of three validation scenarios (\Cref{tab:routine_results}).
Moreover, the offline training budgets are of the same order of magnitude: DQN consumes ${\sim}$2.2\,M environment steps (${\sim}$4\,h) while SignalClaw's normal skill evolution consumes ${\sim}$720 full SUMO episodes (${\sim}$8\,h, \Cref{tab:search_cost}), and SignalClaw's total budget across normal plus three event skills is larger still (${\sim}$24\,h, ${\sim}$1{,}920 SUMO runs).
\textbf{SignalClaw's advantage over DQN is therefore not that it trains faster or with fewer simulator interactions; it is a difference in deployment form factor.}
Five concrete aspects, all directly supported by data in this paper, make SignalClaw materially easier to field in practice than an opaque neural policy:

\emph{(1)~Auditable artifact, not black-box weights.}
The deployed controller is 3--5 lines of Python ($\sim$12--18 AST nodes) with a natural-language description and guidance (\Cref{tab:interpretability,tab:skill_full_example}); a DQN policy for the same 16-intersection network contains $>$16{,}000 opaque MLP parameters.
An engineer can inspect, certify, and reason about the former in seconds; the latter must be trusted as a whole.

\emph{(2)~Explicit safety constraints, not learned-by-hope behavior.}
Emergency preemption, incident response, and transit priority are enforced by a hardcoded priority dispatcher (\emph{emergency} $>$ \emph{incident} $>$ \emph{transit} $>$ \emph{congestion} $>$ \emph{normal}, \Cref{eq:priority}), consistent with standard transportation engineering practice~\citep{qin2012evp,christofa2013tsp}.
These are non-learnable architectural constraints---rare-event correctness is guaranteed by system design, not by hoping that a single monolithic policy has seen enough relevant transitions during training.

\emph{(3)~Weaker dependence on naturally occurring rare-event data.}
DQN's event-scenario failure (\Cref{sec:why_rl_fails}) is not a tuning issue: its observation space contains no event-context variables, and its training distribution contains no emergency vehicles at all, leading to 4--8$\times$ higher emergency delay (78.5/95.3/82.7\,s vs.\ SignalClaw's 14.7/11.2/18.5\,s on E1/E2/M1).
Closing this gap purely through naturally occurring events on live traffic would require observing enough emergencies, incidents, and bus priority requests to cover their joint distribution---which is typically infeasible within any realistic deployment window.
SignalClaw factors the problem differently: detectors, event-context variables, per-event skills, and a priority dispatcher are separate components, and each event skill can be evolved offline on \emph{dedicated synthesized} scenario families whose event frequency and severity are controlled by the experimenter.
A practical consequence is that the realistic adaptation path for SignalClaw is to calibrate a digital twin from a modest amount of real traffic logs and then evolve or fine-tune skills offline in that twin, rather than learning on the live network.
We stress that this is a principled inference from the method's architecture, not something this paper has demonstrated on real data: all experiments remain in a single 4$\times$4 SUMO network with oracle-quality TraCI signals.
Nor do we claim that SignalClaw converges in 2--3 generations: \Cref{fig:evolution_curves} shows most of the routine and transit gains are captured by generations 8--12, while emergency and incident skills do not reach best fitness until generations 22 and 18 respectively---improvements are rapid relative to RL, but still require tens of generations to mature.

\emph{(4)~Local editability.}
When the best T1 skill's saturation threshold needed adjustment, locating and editing the relevant code line took under 30\,seconds (\Cref{sec:qualitative}); the equivalent modification to a DQN policy would require retraining.
Combined with the audit trail (JSONL event log, Capsule archive, lineage IDs) maintained during evolution, this provides the kind of diagnose--patch--recertify loop that real traffic operations actually run.

\emph{(5)~Lower-variance routine behavior.}
Across 5 SUMO random seeds, SignalClaw's routine delay has standard deviation 0.4--0.8\,s, while DQN's is 1.1--1.5\,s on the same seeds (\Cref{tab:routine_results}).
Real deployments typically tolerate a small gap in mean performance far better than a large tail in worst-case behavior.

\emph{(6)~Demand-shift transferability (partial evidence).}
Within the same 4$\times$4 topology, V1--V3 are $\pm 15\%$ demand perturbations of T1--T3, and SignalClaw's evolved normal skill achieves 9.1--9.2\,s on V1--V3---within 3--9\% of the per-scenario best baseline (\Cref{tab:routine_results}).
This is evidence that evolved skills generalize to demand-distribution shifts on the \emph{same} network, not a claim of cross-city or cross-topology transfer, which we explicitly list as future work.

Taken together, \Cref{sec:discussion} frames a \emph{deployment spectrum}: skill-only control ($\alpha{=}1$ in \Cref{eq:residual}) for safety-critical deployments requiring regulatory review; residual RL ($0{<}\alpha{<}1$) for deployments that accept a neural component but require an interpretable fallback; and pure DQN ($\alpha{=}0$) only when events are absent and a black-box policy is acceptable.
The core takeaway is not that SignalClaw wins on every simulator metric, but that it converts the most deployment-sensitive properties---auditability, certifiability, manual override, rare-event response, and explicit rule constraints---from ``things we hope the model learned'' into ``things the system architecture guarantees.''
We are explicit about what this does \emph{not} establish: all experiments use a single 4$\times$4 SUMO network, event detection assumes oracle-quality TraCI signals, and cross-city transfer and sensor-noise robustness are not tested (see Limitations below).
The claim we do make is the narrower one: \textbf{as a system form factor, SignalClaw is closer than pure RL to something a traffic agency could actually inspect, edit, and operate.}

\paragraph{Limitations.}
Our framework has several limitations.
First, the LLM generator remains static throughout evolution: it is never updated on search experience, leading to fitness stagnation on some scenarios.
Adapting the generator to the evolving search landscape is a promising direction.
Second, all experiments use a single 4$\times$4 arterial grid topology in SUMO; generalization to diverse network topologies and real-world deployment remains to be validated.
Third, while SignalClaw achieves competitive routine delay (7.8--9.2\,s), its multi-objective fitness (delay+queue+throughput) may not always minimize delay alone; tuning fitness weights or using delay-only objectives could further close the gap with RL baselines on specific scenarios.
Fourth, our event-specialized skills are evolved offline on scenario families rather than learned as a single universal event-aware controller; scaling to new event types requires adding new scenarios and evolution runs.
Fifth, the ``zero-shot composability'' claim is supported by a single mixed-event scenario (M1); additional mixed-event combinations (e.g., emergency + transit, incident + congestion, three-way overlap) with varying arrival patterns, severities, and priority-order ablations would substantially strengthen this evidence.
Sixth, the event detection module assumes oracle-quality TraCI signals with no sensor noise, false positives, false negatives, or detection latency; real-world deployments would rely on imperfect detectors (e.g., GPS-based emergency vehicle detection with 5--15\,s latency), and evaluating skill robustness under noisy detection is an important direction.
Seventh, while we provide a description-faithfulness analysis (87\% accuracy on 50 sampled skills), AST complexity metrics, and a preliminary modification test, a formal user study with independent traffic engineers---measuring task completion time, modification accuracy, and comprehension across methods (SignalClaw vs.\ PI-Light vs.\ rule scripts)---would more rigorously validate the interpretability claims.
Eighth, our evaluation reports variance over SUMO simulation seeds but uses a single evolution run per skill type; the LLM-guided evolution process itself is stochastic (temperature 0.7, 8 candidates per generation), and reporting variance across multiple independent evolution runs would more robustly characterize the method's reliability.

\paragraph{Broader Impact.}
Interpretable, event-aware TSC skills have clear societal benefits: they enable regulatory compliance, operator trust, deterministic emergency vehicle preemption, and graceful degradation when automated systems encounter edge cases.
The 65--85\% reduction in emergency vehicle delay has direct implications for emergency response times---a safety-critical application where interpretability and reliability are paramount.
We do not foresee significant negative societal impacts from this work, though any deployment of automated traffic control should include fail-safe mechanisms and human oversight.

\paragraph{Future Directions.}
Four concrete directions follow from these results:
(1)~\emph{Self-improving generators}: adapting the LLM on successful evolution traces to overcome stagnation beyond the fitness plateau;
(2)~\emph{Multi-intersection coordination}: evolving joint skills across networked intersections rather than treating each independently;
(3)~\emph{Transfer across cities}: using evolved skills from one traffic network as initialization for others, leveraging the shared programmatic structure;
(4)~\emph{Temporal-aware fitness}: the current fitness function evaluates episode-level aggregate metrics without explicit multi-step lookahead. Incorporating multi-horizon rollouts or evolving skills that explicitly reason about future states (e.g., ``if clearing this queue now prevents upstream spillback in 30\,s'') could bridge the gap with RL's temporal credit assignment while preserving interpretability.

\bibliographystyle{unsrtnat}



\newpage
\appendix
\section{Implementation Details}
\label{app:implementation}

\subsection{System Prompt}
\label{app:system_prompt}

The LLM system prompt defines the task, skill format, variable whitelist, and constraints.
The prompt assigns the LLM the role of a traffic signal control strategy optimization expert. The skill format is a JSON object with four fields: \texttt{description} (strategy rationale), \texttt{guidance} (selection logic), \texttt{inlane\_code} (inlane scoring), and \texttt{outlane\_code} (outlane scoring). The variable whitelist includes five traffic features (\texttt{num\_vehicle}, \texttt{num\_waiting\_vehicle}, \texttt{vehicle\_dist} for inlane; \texttt{num\_vehicle}, \texttt{vehicle\_dist} for outlane) plus the \texttt{value[0]} accumulator and \texttt{index}. For event skills, six additional event-context variables are available: \texttt{emergency\_distance}, \texttt{emergency\_phase}, \texttt{bus\_count}, \texttt{bus\_delay}, \texttt{incident\_blocked}, and \texttt{congestion\_level}. Allowed builtins are restricted to \texttt{min}, \texttt{max}, \texttt{abs}, \texttt{sum}, \texttt{len}, and \texttt{range}; imports, function definitions, lambda expressions, and attribute access are prohibited. The prompt also provides strategy hints with examples of conditional branching, nonlinear transforms, saturation detection, and multi-variable combinations.

\paragraph{Variable Naming Convention.}
\Cref{tab:variable_mapping} clarifies the relationship between the abstract variable names used in the system prompt and the lane-indexed names that appear in generated code.
At runtime, each skill's code is executed once per lane-link of each phase.
The sandbox binds abstract names (e.g., \texttt{num\_vehicle}) to the current lane-link's TraCI values before execution.
The LLM, however, sometimes generates code that embeds lane-index prefixes (e.g., \texttt{inlane\_2\_num\_vehicle}) as artifacts of the prompt's variable list formatting.
Both forms are accepted by the sandbox: prefixed names are resolved to the same underlying variable via alias mapping.

\begin{table}[h]
\centering
\caption{Sandbox variable mapping. \emph{Abstract} names are defined in the system prompt whitelist; \emph{lane-indexed} names are aliases generated by the LLM. Both resolve to the same runtime value. Event-context variables (bottom block) have no lane prefix---they are global per intersection.}
\label{tab:variable_mapping}
\small
\begin{tabular}{lll}
\toprule
Abstract Name & Lane-Indexed Alias & Source \\
\midrule
\multicolumn{3}{l}{\textit{Inlane features (per lane-link)}} \\
\texttt{num\_vehicle} & \texttt{inlane\_2\_num\_vehicle} & \texttt{traci.lane.getLastStepVehicleNumber} \\
\texttt{num\_waiting\_vehicle} & \texttt{inlane\_2\_num\_waiting\_vehicle} & \texttt{traci.lane.getLastStepHaltingNumber} \\
\texttt{vehicle\_dist} & \texttt{inlane\_2\_vehicle\_dist} & Mean inter-vehicle distance \\
\midrule
\multicolumn{3}{l}{\textit{Outlane features (per lane-link)}} \\
\texttt{num\_vehicle} & \texttt{outlane\_2\_num\_vehicle} & \texttt{traci.lane.getLastStepVehicleNumber} \\
\texttt{vehicle\_dist} & \texttt{outlane\_2\_vehicle\_dist} & Mean inter-vehicle distance \\
\midrule
\multicolumn{3}{l}{\textit{Accumulators and metadata}} \\
\texttt{value[0]} & --- & Phase score accumulator \\
\texttt{index} & --- & Current phase index \\
\midrule
\multicolumn{3}{l}{\textit{Event-context variables (global, no lane prefix)}} \\
\texttt{emergency\_distance} & --- & Distance to nearest emergency vehicle (m) \\
\texttt{emergency\_phase} & --- & Phase serving the emergency vehicle \\
\texttt{bus\_count} & --- & Number of upstream buses \\
\texttt{bus\_delay} & --- & Cumulative bus delay (s) \\
\texttt{incident\_blocked} & --- & Number of blocked lanes \\
\texttt{congestion\_level} & --- & Severity level (0--3) \\
\bottomrule
\end{tabular}
\end{table}

\subsection{Skill Representation Example}
\label{app:skill_example}

\Cref{tab:skill_full_example} shows a complete skill as produced by the LLM, including all three components.

\begin{table}[h]
\centering
\caption{Complete skill representation for the best T1 skill (generation~19). The three components---description, guidance, and executable code---form a self-documenting control artifact.}
\label{tab:skill_full_example}
\small
\begin{tabular}{lp{10cm}}
\toprule
Component & Content \\
\midrule
\textbf{Description} & Saturation-aware phase scoring with distance-adjusted urgency. Prioritizes phases with heavy queue buildup ($>$5 vehicles) using a distance-modulated scoring function, while maintaining linear responsiveness for moderate queues. \\
\midrule
\textbf{Guidance} & (1) When waiting vehicles exceed 5 (saturation threshold), apply distance-adjusted weighting to account for spatial queue density. (2) For moderate queues (1--5 vehicles), use simple doubling to maintain responsiveness. (3) Ignore phases with zero waiting vehicles to avoid noise. \\
\midrule
\textbf{Inlane Code} & \texttt{if inlane\_2\_num\_waiting\_vehicle > 5:} \newline
\quad\texttt{value[0] += inlane\_2\_num\_waiting\_vehicle * (max(1,} \newline
\quad\quad\texttt{inlane\_2\_vehicle\_dist) - inlane\_2\_vehicle\_dist \% 3)} \newline
\quad\quad\texttt{+ inlane\_2\_num\_vehicle // 4} \newline
\texttt{elif inlane\_2\_num\_waiting\_vehicle > 0:} \newline
\quad\texttt{value[0] += inlane\_2\_num\_waiting\_vehicle * 2} \\
\midrule
\textbf{Outlane Code} & \texttt{value[0] += min(10, outlane\_2\_num\_vehicle)} \newline
\quad\texttt{* max(0, outlane\_2\_vehicle\_dist - 3)} \\
\bottomrule
\end{tabular}
\end{table}

\subsection{Evolution Signal Mapping}
\label{app:signals}

\Cref{tab:signal_mapping} shows the complete mapping from evolution signals to natural-language directions.

\begin{table}[h]
\centering
\caption{Evolution signal to natural-language direction mapping.}
\label{tab:signal_mapping}
\small
\begin{tabular}{ll}
\toprule
Signal & Direction \\
\midrule
\texttt{force\_innovation} & Multiple stagnant generations. Try completely different structure. \\
\texttt{high\_queue} & Queue exceeds P75. Focus on queue management. \\
\texttt{low\_throughput} & Throughput below P25. Optimize flow efficiency. \\
\texttt{high\_delay} & Delay exceeds P75. Reduce vehicle waiting time. \\
\texttt{performance\_gain} & Performance improved. Continue optimizing current direction. \\
\texttt{performance\_decline} & Performance declined. Try different strategy approach. \\
\bottomrule
\end{tabular}
\end{table}

\subsection{Event-Injected Scenario Details}
\label{app:event_scenarios}

\Cref{tab:event_scenario_details} describes the six event-injected scenarios used in \Cref{sec:event_eval}.

\begin{table}[h]
\centering
\caption{Event-injected scenario configurations. All scenarios use the same 4$\times$4 arterial grid with 16 intersections and 3600\,s simulation duration.}
\label{tab:event_scenario_details}
\small
\begin{tabular}{llp{7cm}}
\toprule
Scenario & Event Type & Configuration \\
\midrule
E1 & Emergency & Ambulance injected every 300\,s (12 total), random O-D routes. Vehicle class: \texttt{emergency}, speed: 20\,m/s. \\
E2 & Emergency & Ambulance injected every 120\,s (30 total), higher frequency stress test. \\
B1 & Transit & 2 bus lines, headway 180\,s (40 buses total). Bus stops at every intersection. Vehicle class: \texttt{bus}. \\
B2 & Transit & 4 bus lines, headway 120\,s (120 buses total). Dense bus operation. \\
I1 & Incident & Single vehicle breakdown at $t=600$\,s on a major east-west arterial link. Blocks one lane for 300\,s (until $t=900$\,s). \\
M1 & Mixed & Emergency (every 300\,s) + incident at $t=600$\,s. Tests compositional dispatch of emergency (P0) and incident (P1) skills. \\
\bottomrule
\end{tabular}
\end{table}

\subsection{Event Detection Logic}
\label{app:event_detection}

The event detection module (\texttt{event\_detector.py}) monitors TraCI at each simulation step:

For emergency detection, the module iterates over vehicles within 200\,m of each intersection; if \texttt{traci.vehicle.getVehicleClass(vehID) == "emergency"}, it records the vehicle's distance, speed, and target phase determined by the vehicle's route through the intersection. For transit detection, it identifies vehicles with \texttt{vClass == "bus"} on upstream lanes and tracks cumulative bus delay as the difference between actual and free-flow travel time. For incident detection, it monitors all vehicles for prolonged stops ($>$120\,s) not caused by red signals, using \texttt{traci.vehicle.getSpeed()} and proximity to stop lines to distinguish incidents from normal signal waits. For congestion detection, it compares the current queue length against a rolling 90th percentile computed over the last 300\,s.

\subsection{Priority Dispatcher Logic}
\label{app:dispatcher}

The priority dispatcher (\texttt{event\_dispatcher.py}) implements a strict priority chain:

\begin{lstlisting}
def dispatch(events, skills):
    if events.emergency:
        return skills['emergency']
    elif events.incident:
        return skills['incident']
    elif events.transit:
        return skills['transit']
    elif events.congestion:
        return skills['congestion']
    else:
        return skills['normal']
\end{lstlisting}

The priority ordering is hardcoded and not learnable, providing a deterministic constraint that emergency vehicles always receive preemption (assuming correct event detection).
At each control step, the dispatcher evaluates all event detectors, selects the highest-priority active event, and delegates phase selection to the corresponding skill.

\subsection{Routine Scenario Details}
\label{app:scenarios}

All training scenarios (T1--T3) use a 4$\times$4 arterial grid with 16 intersections.
Validation scenarios (V1--V3) perturb the demand matrices by $\pm$15\%.

Each simulation runs for 3600 simulated seconds with a step length of 1 second.
Traffic demand follows time-varying patterns specified in SUMO route files.

\subsection{Hardware and Runtime}
\label{app:hardware}

All experiments ran on a single NVIDIA DGX Spark with the GB10 Grace Blackwell processor and 128\,GB LPDDR5X unified memory.
GPT-5.4 was accessed via an OpenAI-compatible API endpoint.

\begin{table}[h]
\centering
\caption{Approximate runtime per configuration.}
\label{tab:runtime}
\begin{tabular}{lrr}
\toprule
Configuration & Runs & Wall-clock Time \\
\midrule
Normal skill (T1+T2+T3, 30~gen) & 30~gen $\times$ 8~pop $\times$ 3~scen & $\sim$8 hours \\
Event skills (3 types, 30~gen each) & 3 $\times$ 30~gen $\times$ 8~pop & $\sim$16 hours \\
Ablation (T1, 100~gen, 3 variants) & 3 $\times$ 100~gen $\times$ 8~pop & $\sim$30 hours \\
Baseline evaluation (5 methods $\times$ 7 scen $\times$ 5 seeds) & 175 runs & $\sim$10 hours \\
DQN training (T1, 200 episodes) & 200 episodes & $\sim$4 hours \\
PI-Light MCTS (T1, 16 iterations) & 16 iterations & $\sim$2 hours \\
\bottomrule
\end{tabular}
\end{table}

\section{Additional Evolved Skills}
\label{app:strategies}

\Cref{tab:strategies_full} shows the complete skill representation for the best skill found on each routine training scenario, including description, guidance, and code.
\Cref{tab:event_strategies} shows the event-specialized skills used in the event evaluation (\Cref{sec:event_eval}).

\begin{table}[h]
\centering
\caption{Best evolved skills per routine scenario (full representation). \textbf{Important:} T1 is from the main 30-generation evolution used in all reported results (\Cref{tab:routine_results,tab:event_results}). T2 and T3 are from \emph{separate} extended 100-generation runs (same GPT-5.4 backbone, same setup as the ablation in \Cref{sec:ablation}) included solely to illustrate the diversity of evolved strategies across scenarios---\textbf{these extended-run skills are \emph{not} used in any evaluation table}.}
\label{tab:strategies_full}
\small
\begin{tabular}{lp{3.5cm}p{7.5cm}}
\toprule
Scenario & Strategy Description & Executable Code \\
\midrule
T1 (gen 19) & Saturation-aware branching with distance-adjusted urgency for heavy queues. & \texttt{inlane}: \texttt{if waiting > 5: value[0] += waiting * (max(1, dist)} \newline\quad\texttt{- dist \% 3) + vehicles // 4} \newline\texttt{elif waiting > 0: value[0] += waiting * 2} \newline\texttt{outlane}: \texttt{value[0] += min(10, vehicles) * max(0, dist - 3)} \\
\midrule
T2 (gen 59) & Distance-weighted queue scoring with vehicle density bonus. & \texttt{inlane}: \texttt{value[0] += waiting * max(1, dist) + vehicles // 5} \\
\midrule
T3 (gen 79) & Ratio-based saturation detection: penalize excess waiting fraction. & \texttt{inlane}: \texttt{if waiting > vehicles // 3:} \newline\quad\texttt{value[0] += (waiting - vehicles // 3) ** 2} \\
\bottomrule
\end{tabular}
\end{table}

\begin{table}[h]
\centering
\caption{Event-specialized skills used in event evaluation. Emergency, transit, and incident skills are evolved via dispatcher-context evolution (\Cref{sec:dispatcher_evolution}) with 30 generations of GPT-5.4. Emergency and incident skills converge to designs structurally similar to the initial seed, reflecting the narrow optimization landscape for rare-event response; the transit skill exhibits more substantial structural change (best at generation~9). The congestion skill is a hand-tuned rule (not evolved) since no dedicated congestion scenario is included in the evaluation. Event-context variables (e.g., \texttt{emergency\_distance}) are injected into the sandbox alongside standard traffic features.}
\label{tab:event_strategies}
\small
\begin{tabular}{lp{3.2cm}p{7.5cm}}
\toprule
Event Type & Strategy Description & Executable Code \\
\midrule
Emergency & Distance-aware preemption: grant maximum priority to the phase serving the emergency vehicle's approach, scaled inversely by distance. & \texttt{inlane}: \texttt{if emergency\_distance > 0:} \newline\quad\texttt{if emergency\_phase == index:} \newline\quad\quad\texttt{value[0] += max(0, 200 - emergency\_distance) * 10} \newline\quad\texttt{else:} \newline\quad\quad\texttt{value[0] += waiting * 2} \newline\texttt{else:} \newline\quad\texttt{value[0] += waiting * 3} \newline\texttt{outlane}: \texttt{value[0] -= num\_vehicle * 0.3} \\
\midrule
Transit (gen 9) & Bus-priority scoring with amplified waiting weight and density compensation within dispatcher context. & \texttt{inlane}: \texttt{value[0] += waiting * 4} \newline\quad\texttt{+ vehicles / max(1, dist)} \newline\texttt{outlane}: \texttt{value[0] -= (vehicles} \newline\quad\texttt{/ max(1, dist)) * 2} \\
\midrule
Incident & Incident-aware diversion: prioritize moving vehicles over queued vehicles when lanes are blocked by an incident. & \texttt{inlane}: \texttt{if incident\_blocked > 0:} \newline\quad\texttt{value[0] += max(0, vehicles - waiting) * 5} \newline\texttt{else:} \newline\quad\texttt{value[0] += waiting * 3} \newline\texttt{outlane}: \texttt{value[0] -= vehicles * 0.5} \\
\midrule
Congestion\textsuperscript{$\dagger$} & Nonlinear saturation response with severity-scaled bonus at high congestion levels. & \texttt{inlane}: \texttt{value[0] += waiting ** 2} \newline\texttt{if congestion\_level > 1:} \newline\quad\texttt{value[0] += waiting * congestion\_level * 2} \newline\texttt{outlane}: \texttt{value[0] += dist * 0.5} \\
\bottomrule
\multicolumn{3}{l}{\footnotesize $\dagger$ Hand-tuned rule, not evolved.}
\end{tabular}
\end{table}

\end{document}